\documentclass[11pt]{article}
\usepackage{acl}
\usepackage{times}
\usepackage{latexsym}
\usepackage[T1]{fontenc}
\usepackage[utf8]{inputenc}
\usepackage{graphicx}
\usepackage{amsmath,amssymb,amsfonts}
\usepackage{booktabs}
\usepackage{multirow}
\usepackage{enumitem}
\usepackage{microtype}
\usepackage{xcolor}

\newcommand{\KD}{\textsc{KD}}
\newcommand{\RCB}{\textsc{RCB}}
\newcommand{\OT}{\textsc{OT}}

\title{Decodable but Not Corrected by Fixed Residual-Stream Linear Steering:\\Evidence from Medical LLM Failure Regimes}

\author{Ming Liu \\
  Amazon \\
  \texttt{mlliuz@amazon.com}}

\begin{document}
\maketitle

\begin{abstract}
Can linearly decodable failure signals in LLM hidden states be leveraged to correct those failures? We investigate this \emph{classification-correction gap} via \emph{Overthinking} (\OT{})---a stable behavioral regime (Jaccard $\geq 0.81$, 94\% inter-annotator agreement) in medical QA where models answer correctly under resampling yet fail in extended chain-of-thought. \OT{} is linearly decodable at 71.6\% balanced accuracy ($p \approx 10^{-16}$). Yet five families of fixed linear steering (29 configurations, $n = 1{,}273$) all yield $\Delta \approx 0$, with identical null results cross-architecture (Qwen2.5-7B) and cross-domain (MMLU-STEM). Three convergent lines of evidence suggest \emph{representational entanglement}: the \OT{} direction has 85--88\% overlap with task-critical computation (specificity ratio $\leq 0.152$); non-targeted shared-direction steering damages accuracy ($-$12.1pp); and LEACE concept erasure damages accuracy ($-$3.6pp, $p = 0.01$), while 10 random erasures produce $\Delta = +0.3$pp. The per-instance probe--steering correlation is $r = -0.002$ ($p = 0.97$). Positively, the same probe enables selective abstention (held-out AUROC = 0.610, exceeding all five uncertainty baselines, $p = 0.009$): decodable failure structure supports post-generation reliability estimation even when the fixed linear steering family cannot exploit it for correction.
\end{abstract}

\section{Introduction}

Activation steering---adding directions in activation space to modify model behavior---succeeds for binary behavioral properties like refusal \citep{arditi2024refusal} and sentiment \citep{turner2024activation}. But does this extend to correcting multi-step reasoning failures? Chain-of-thought prompting \citep{wei2022chain,kojima2022zeroshot} enables multi-step reasoning but also introduces failure modes tied to extended generation \citep{turpin2023language,lanham2023measuring}; indeed, \citet{huang2024selfcorrect} show that LLMs cannot self-correct reasoning without external feedback. We use medical question answering as a controlled testbed: it demands multi-step reasoning, failures carry distinct semantic signatures, and the multiple-choice format provides an unambiguous evaluation endpoint.

We focus on a specific empirical question: \emph{when models frequently answer correctly yet sometimes fail through extended reasoning, can the representational structure underlying this behavioral regime be leveraged for activation steering?}

We address this question through a systematic investigation with four contributions:

\begin{enumerate}[leftmargin=*,itemsep=2pt]
    \item \textbf{Behavioral regime identification.} We identify \emph{Overthinking} (\OT{})---where models frequently answer correctly under alternative sampled traces but produce incorrect long-reasoning outputs---as a robust behavioral construct, stable under threshold perturbation, length regression, and prompt-end probing (Section~\ref{sec:taxonomy}).

    \item \textbf{Geometric analysis.} Binary \OT{}-vs-non-\OT{} classification achieves 71.6\% (modest but highly reliable; $p \approx 10^{-16}$), with supporting evidence on Qwen2.5-7B and cross-domain transfer to MMLU-STEM (decodability and steering failure). The OT-specific specificity ratio is only 0.119 (88\% of the signal is shared with task-relevant directions), and concept erasure damages accuracy ($-3.6$pp, $p = 0.01$), providing causal evidence that the decodable signal is entangled with task computation rather than merely correlational.

    \item \textbf{An empirical classification-correction gap.} Five families of fixed linear steering (contrastive, probe-guided, multi-layer, prompt-end, and subspace; 29 configurations total, $n = 1{,}273$) on Llama-3.1-8B produce $\Delta \approx 0$, while shared-direction steering ($-$12.1pp) and mean-difference concept erasure ($-$3.6pp, $p = 0.01$; direction-specific vs.\ random erasures) damage accuracy---consistent with failure-mode information co-occurring with task-relevant computation. Qwen2.5-7B (9 configurations), MMLU-STEM (4 configurations), and ten prompt baselines including self-refinement and verbalized confidence are consistent with this pattern. A refusal-steering control verifies the implementation ($p = 0.008$, one-sided).

    \item \textbf{Post-generation reliability estimation.} The same structure enables selective abstention after a single generation: a correctness probe (layer 21, selected via held-out AUROC from a 32-layer scan) achieves held-out split AUROC = 0.716 (held-out test: 0.610; generalization gap is primarily temperature-driven), outperforming all five tested single-forward-pass uncertainty baselines ($\Delta$AUROC = 0.041, $p = 0.009$; Holm rank-1 threshold = 0.010). \emph{Reading} representations has operational value even when linear \emph{writing} to them does not---probing success, even when robust and cross-domain transferable, does not imply steerability.
\end{enumerate}

\section{Related Work}

\paragraph{Linear directions can steer binary behavioral attributes---but unreliably.}
Representation engineering \citep{zou2023representation} shows that linear directions correspond to interpretable concepts. Inference-time intervention \citep{li2024inference}, activation addition \citep{turner2024activation}, contrastive activation addition \citep{panickssery2023steering}, mean-centred variants \citep{jorgensen2023meancentring}, and feature clamping via sparse autoencoders \citep{templeton2024scaling,cunningham2023sparse} demonstrate successful steering for binary behavioral traits (refusal, sentiment, sycophancy \citep{sharma2024sycophancy}) and in-context task identity \citep{todd2024function}, while model editing \citep{meng2022locating} modifies localized factual associations. Layer-contrast decoding \citep{chuang2024dola} provides an alternative inference-time intervention that improves factuality by exploiting layer-wise knowledge maturation. Representation finetuning \citep{wu2024reft} instead \emph{learns} targeted interventions on frozen representations. However, growing evidence reveals systematic steering failures: \citet{silva2025steering} find substantial variability across 36 models from 14 families; \citet{braun2025sober} document concepts that remain unsteerable across all layers; \citet{mckenzie2026endogenous} identify endogenous resistance circuits that actively counteract interventions; and \citet{zur2025road} show steering fails once a model has committed to a reasoning path---even when hidden states still encode uncertainty. \citet{jafari2026mechanistic} propose mechanistic indicators for predicting effectiveness a priori, and \citet{billa2026predicting} formalize a three-regime framework in which some concepts are provably beyond any linear intervention's reach.

\paragraph{But probed information need not be causally effective.}
The linear representation hypothesis \citep{park2024linear,marks2024geometry,nanda2023emergent,hernandez2024linearity} and probing methods \citep{alain2017understanding,belinkov2022probing,hewitt2019control} establish that LLMs encode rich structure---including correctness signals \citep{azaria2023internal,burns2023discovering}---but whether probed information is \emph{causally} used remains contested \citep{elazar2021amnesic,ravfogel2022linear}. \citet{deng2025rethinking} argue confounding bias can cause probes to find correlational rather than causal directions; however, our LEACE erasure result ($-3.6$pp damage, direction-specific) rules out the purely correlational account in our setting, instead supporting causal entanglement with task computation. Concept erasure methods---INLP \citep{ravfogel2020null}, RLACE \citep{ravfogel2022linear}, LEACE \citep{belrose2023leace}---and causal mediation \citep{vig2020investigating,geiger2024finding} test this distinction. \citet{hase2024localization} show that localizing facts does not predict editing success, foreshadowing the gap we observe.

The decodability--steerability gap is now documented across multiple domains. \citet{wu2025axbench} benchmark this at scale: on Gemma-2, difference-of-means probes achieve best concept detection while prompting outperforms all representation-based steering methods. \citet{basu2026interpretability} find 98\% AUROC internal hazard representations in clinical triage that four mechanistic interventions fail to exploit. \citet{sanyal2025confidence} and \citet{cox2026decoding} report analogous gaps in math solvability and factual reasoning. \citet{wang2026asa} coin ``representation-behavior gap'' for the same phenomenon in tool-calling agents (99\% probe AUC, model still fails to act). \citet{mishra2026nonsurjective} prove steered activations leave the prompt-reachable set; \citet{nadaf2026steerable} document the inverse (steerability without decodability). Theoretically, \citet{gao2026cylindrical} extend the linear representation hypothesis to show that concept overlap creates intrinsically unpredictable ``sensitive sectors,'' while \citet{wollschlager2025geometry} prove that geometric orthogonality does not guarantee interventional independence. When representations are encoded in superposition \citep{elhage2022superposition}---sharing directions across features---additive steering along one feature's direction inevitably perturbs others.

\paragraph{Our contribution: mechanistic explanation, not just documentation.}
While the gap's existence is now established across several concurrent works \citep{wu2025axbench,basu2026interpretability,wang2026asa}, none provides: (a)~systematic geometric characterization of \emph{why} steering fails in a specific domain---we show the OT-specific specificity ratio is only 0.119, meaning 88\% of the contrastive signal is shared with task-relevant computation, and concept erasure \emph{damages} accuracy ($-3.6$pp, $p = 0.01$), confirming the entanglement is causal rather than merely correlational; (b)~per-instance analysis showing probe confidence and steering effect are completely uncorrelated ($r = -0.002$, $p = 0.97$); or (c)~an operational positive---redirecting decodable structure toward selective abstention. We test the commonly assumed probing-to-steering move in multi-step medical reasoning \citep{jin2021disease,singhal2023large,nori2023generalist}---a domain where overthinking degrades accuracy \citep{chen2024overthinking,wang2025underthinking} and chain-of-thought benefits remain limited compared to symbolic tasks \citep{sprague2024cot}---with five method families (29 configurations), and provide convergent geometric evidence (low specificity, directional damage, cross-domain subspace degradation) that the gap arises from representational entanglement with task computation, consistent with the structurally unreachable regime of \citet{billa2026predicting}. This contrasts with domains where steering succeeds (factual recall \citep{li2024inference}, refusal \citep{arditi2024refusal})---where target directions have higher specificity ratios (refusal: 0.999 vs.\ \OT{}: $\leq$0.21; a $4.7\times$ gap that is predictive of steerability in our two-point comparison).

\section{Failure Characterization}
\label{sec:taxonomy}

We characterize failures along two axes. \textbf{Axis A (primary): Behavioral regime} (\OT{} vs.\ non-\OT{}), defined by cross-trace statistics and carrying our strongest claims. \textbf{Axis B (exploratory): Error mechanism} (\KD{} vs.\ \RCB{}, within non-\OT{} only), defined by semantic judgment with lower inter-annotator agreement. Core findings do not depend on the \KD{}/\RCB{} boundary.

\paragraph{Primary construct: Overthinking (\OT{}).} We use ``Overthinking'' as a behavioral label for a sampling-level pattern, not as a claim about internal cognitive mechanism; cf.\ \citet{chen2024overthinking} on excessive computation in o1-like models and \citet{muennighoff2025s1} on budget forcing. Operationally: the model answers correctly in $\geq$60\% of sampled traces \citep{wang2023selfconsistency}---a conservative majority threshold---but produces incorrect answers in long-reasoning traces ($>$200 tokens). The length threshold is nearly redundant ($>$99\% of incorrect traces exceed 200 tokens), so the effective \OT{} boundary is determined by the correct-rate threshold. \OT{} is defined by \emph{observable behavioral statistics}, yielding high reproducibility: 94\% expert agreement, 100\% within-question purity, and Jaccard $\geq$0.81 under threshold perturbation (Section~\ref{sec:geometric}). Length regression and prompt-end probing (54.4\% vs.\ 50\% chance) confirm \OT{} reflects a generation-time regime rather than question difficulty or response length alone. Our research question is whether representational structure tracking this regime can support correction, regardless of what mechanism produces it \citep{turpin2023language}.

\paragraph{Exploratory sub-classification: \KD{} and \RCB{}.} Within non-\OT{} failures, \emph{Knowledge Deficit} (\KD{}) denotes factual errors and \emph{Reasoning Chain Break} (\RCB{}) denotes valid-fact/invalid-logic errors. This distinction requires semantic judgment ($\kappa = 0.61$ expert-expert, $\kappa = 0.30$ LLM; Section~\ref{sec:taxonomy_validation}).

\subsection{Annotation Pipeline}

Annotation proceeds in two phases: (1)~heuristic OT detection---if $\geq$60\% of 10 sampled traces are correct, incorrect traces $>$200 tokens are labeled \OT{}; (2)~LLM-based \KD{}/\RCB{} classification of remaining traces via Claude Haiku \citep{anthropic2024claude} following the LLM-as-judge paradigm \citep{zheng2024judging}. Full prompt and pipeline details are in Appendix~\ref{app:annotation}.

\subsection{Annotation Statistics}

\begin{table}[t]
\centering
\footnotesize
\setlength{\tabcolsep}{3pt}
\begin{tabular}{lrrrr}
\toprule
& \KD{} & \RCB{} & \OT{} & Unclear \\
\midrule
Llama-3.1-8B & 7,419 & 11,895 & 10,713 & 4,505 \\
 & (21.5\%) & (34.5\%) & (31.0\%) & (13.0\%) \\
\midrule
Qwen2.5-7B & 12,960 & 13,629 & 6,306 & 6,967 \\
 & (32.5\%) & (34.2\%) & (15.8\%) & (17.5\%) \\
\bottomrule
\end{tabular}
\caption{Failure mode annotation distribution. Values show count and percentage of all incorrect traces. Llama annotations are over 34,532 traces from 10,178 MedQA questions (10 traces/question). Qwen annotations are over 39,862 traces from the same questions.}
\label{tab:annotations}
\end{table}

Table~\ref{tab:annotations} shows the annotation distribution. \RCB{} is the most common failure mode (34.5\%), followed by \OT{} (31.0\%) and \KD{} (21.5\%), with consistent proportions across architectures. Unclear cases (13.0\%/17.5\%) are excluded from subsequent analysis.

\subsection{Annotation Validation}
\label{sec:taxonomy_validation}

Domain expert validation ($n = 500$ gold set, two blinded 4th-year clinical students) yields 94\% agreement on \OT{}, with expert-expert $\kappa = 0.61$ on the three-way task; the \KD{}/\RCB{} boundary is the primary source of disagreement. Within-question \OT{} purity is 100\% (all incorrect traces from an \OT{} question receive the same label). Simulating 37\% \KD{}/\RCB{} label noise (matching observed LLM disagreement) drops three-way classification only 2.9pp while leaving the binary \OT{}-vs-non-\OT{} result (71.6\%) unaffected. Full validation details (LLM cross-validation, multi-model comparison, within-question consistency analysis) are in Appendix~\ref{app:annotation_validation}.

\subsection{Behavioral Validation}

\OT{} shows a steep negative length-accuracy gradient ($\rho = -0.164$, $p < 0.001$), confirming that extended reasoning is associated with lower accuracy, while \KD{}/\RCB{} show weak slopes ($\rho = -0.044, -0.082$ respectively). \OT{} has 78.3\% base accuracy vs.\ $\sim$28\% for \KD{}/\RCB{}, confirming a clean two-way behavioral split. Self-consistency \citep{wang2023selfconsistency} (majority vote, 10 traces) achieves AUROC = 0.804 for correctness prediction and---by construction of the $\geq$60\% threshold---corrects 100\% of OT-incorrect questions, yielding $+10.6$pp on training and $+7.4$pp overall on the held-out test ($p < 10^{-10}$; Section~\ref{sec:discussion}). A single-forward-pass correctness probe achieves AUROC = 0.716 (held-out 30\% split; 5-fold CV: 0.721 $\pm$ 0.005; independent test set: 0.610) using a geometrically different signal (Section~\ref{sec:geometric}), substantially exceeding surface uncertainty baselines (all $\leq 0.530$ on training, $\leq 0.569$ on test; $\Delta$AUROC = 0.041, $p = 0.009$; Tables~\ref{tab:uncertainty}--\ref{tab:uncertainty_test}).

\section{Geometric Analysis of Failure Modes}
\label{sec:geometric}

\subsection{Hidden State Extraction}

We extract hidden states from Llama-3.1-8B-Instruct \citep{grattafiori2024llama} for 12,000 traces: 2,000 per failure mode (\KD{}, \RCB{}, \OT{}) plus 6,000 matched correct traces (2,000 correct traces from the same questions as each failure mode). For each trace, we construct the full conversation (system prompt + question + model response), extract the last-token hidden state at all 32 layers in fp32 precision, yielding a tensor of shape $(12000, 32, 4096)$.

\subsection{Contrastive Vectors}

Following the representation engineering framework, we compute contrastive vectors:
\begin{equation}
    \mathbf{v}_m^{(l)} = \frac{\bar{\mathbf{h}}_{\text{correct}}^{(l)} - \bar{\mathbf{h}}_m^{(l)}}{\|\bar{\mathbf{h}}_{\text{correct}}^{(l)} - \bar{\mathbf{h}}_m^{(l)}\|}
\end{equation}
where $\bar{\mathbf{h}}_{\text{correct}}^{(l)}$ and $\bar{\mathbf{h}}_m^{(l)}$ are the mean hidden states at layer $l$ for correct traces and mode $m$ traces, respectively.

\begin{figure*}[t]
\centering
\includegraphics[width=\textwidth]{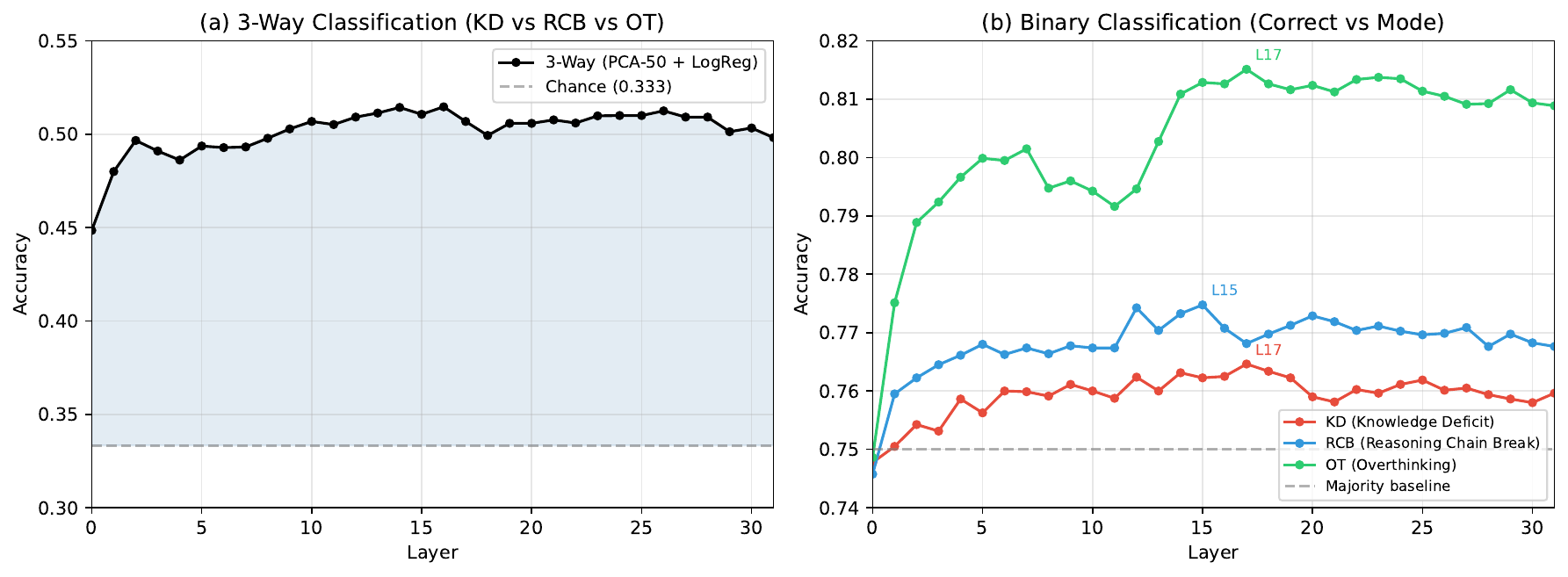}
\caption{Layer-wise classification accuracy profiles. (a) Exploratory three-way classification (\KD{}/\RCB{}/\OT{}) using PCA-50 + logistic regression, peaking at 51.5\% (chance = 33.3\%). (b) Binary classification (correct vs.\ mode) showing mode-specific layer profiles: \OT{} peaks at layer 17 (81.5\%), providing the strongest and most annotation-robust signal.}
\label{fig:layer_classification}
\end{figure*}

\subsection{Statistical Analysis Suite}

\begin{figure}[t]
\centering
\includegraphics[width=\columnwidth]{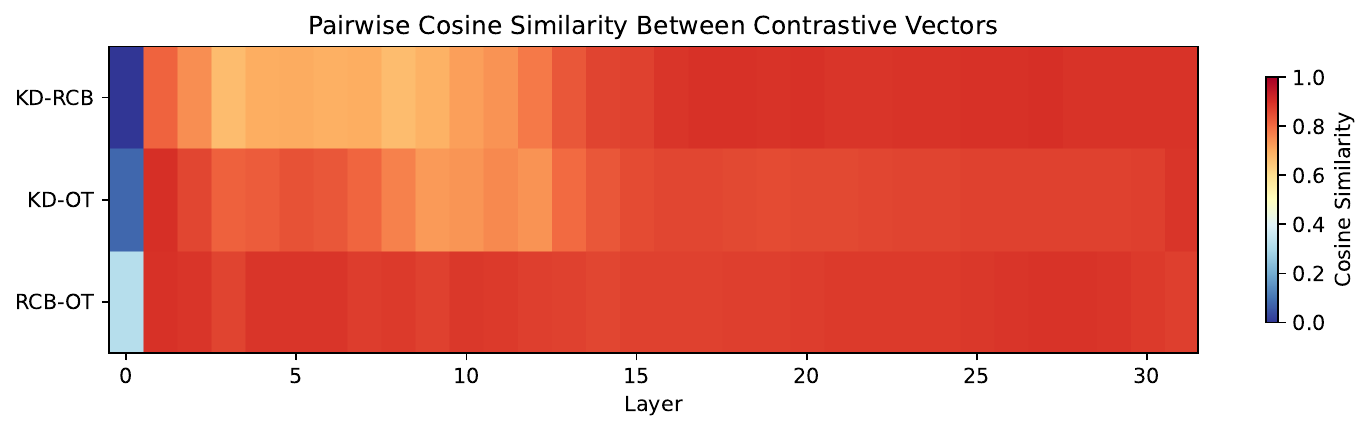}
\caption{Pairwise cosine similarity between contrastive vectors across layers. Early layers show low similarity (mode-specific information), while later layers converge to high similarity ($>$0.9), reflecting the dominance of the shared component.}
\label{fig:cosine_heatmap}
\end{figure}

\paragraph{Classification.}
Three-way \KD{}/\RCB{}/\OT{} classification reaches 51.5\% at layer 16 (chance = 33.3\%; Figure~\ref{fig:layer_classification}a). Per-mode correct-vs-incorrect probes peak at 76--82\% with mode-specific layer profiles (Figure~\ref{fig:layer_classification}b). Binary \OT{}-vs-non-\OT{} classification achieves \textbf{71.6\%} accuracy (balanced accuracy = 62.3\%, AUROC = 0.672; majority baseline = 66.7\%, $p \approx 10^{-16}$). The effect size is modest (4.9pp above majority baseline): the probe achieves high specificity (90\% non-\OT{} recall) but low sensitivity (34\% \OT{} recall), reflecting moderate but statistically reliable separability. Contrastive vectors show high pairwise cosine similarity (mean 0.820; Figure~\ref{fig:cosine_heatmap}), indicating substantial overlap across modes. The observed pairwise distance (3.38) is 2.4$\times$ the permutation null ($p < 0.001$); shuffled labels yield chance-level classification (34.6\%), ruling out data artifacts.

\subsection{Robustness Controls}

The 71.6\% binary classification is robust to six potential confounds (full details in Appendix~\ref{app:robustness_controls}): random labels yield chance (33.2\%); length regression leaves \OT{}-vs-non-\OT{} unaffected (71.3\%); threshold sweeps produce Jaccard $\geq$0.81; GroupKFold rules out question-identity leakage ($<$1pp drop). Prompt-end probing achieves only 54.4\% (vs.\ 71.6\% at last-token), confirming the probe detects a generation-time regime rather than question difficulty.

\subsection{Direction Ablation Analysis}

Removing 3 contrastive directions (from 4096) drops in-sample classification from 51.4\% to 25.8\% (below chance), while removing random directions has zero effect. However, cross-validated ablation (directions fit on split A, evaluated on split B) shows no significant drop ($\Delta = -0.2$pp), indicating that the in-sample effect reflects probe-direction overfitting rather than a uniquely necessary subspace. These directions produce no reliable behavioral change under additive steering (Section~\ref{sec:steering})---necessity for a probe's classification does not entail causal involvement in the model's generation process.

\subsection{Shared-Specific Decomposition}

The average specificity ratio is only \textbf{0.119}: 88\% of each contrastive vector aligns with a shared ``incorrect-vs-correct'' direction, confirmed under binary \OT{}-vs-non-\OT{} framing (bypassing \KD{}/\RCB{} entirely). A permutation test (10,000 label shuffles among incorrect traces) yields null specificity mean = 0.370; observed = 0.119 is significantly \emph{below} all permuted values ($p < 0.0001$), confirming anomalously high alignment with the shared axis rather than an artifact of set-subset structure. To further rule out circularity (since \OT{} traces are a subset of incorrect traces), we compute $\mathbf{d}_{\text{OT vs KD}} = \bar{\mathbf{h}}_{\text{OT}} - \bar{\mathbf{h}}_{\text{KD}}$---a direction that makes no reference to correct traces---and find $\cos(\mathbf{d}_{\text{OT vs KD}}, \mathbf{d}_{\text{correctness}}) = 0.86$, demonstrating that the entanglement persists even when the correct class is removed from the computation. Uniform steering using this shared component is harmful ($\Delta = -12.1$pp; Table~\ref{tab:steering}), consistent with the failure-mode direction sharing representational subspace with task-relevant computation.

\subsection{Probe-Based Analysis}

Probe weight vectors are geometrically dissociated from contrastive vectors (cosine 0.23--0.54 vs.\ 0.82), yet achieve comparable classification (51.2\% vs.\ 51.5\%). This dissociation---same accuracy, different directions---illustrates that decodability does not uniquely identify an intervention target (Appendix~\ref{app:probe_analysis}). Non-linear probes (MLP, SVM-RBF) perform at or below linear on classification tasks ($\Delta \leq -0.2$pp for MLP); SVM-RBF gains +1.4pp on correctness AUROC (0.734 vs.\ 0.720), a marginal improvement confirming the signal ceiling is close to what linear probes capture.

\subsection{Cross-Architecture Validation}

We repeat the full geometric analysis suite on Qwen2.5-7B-Instruct \citep{qwen2024qwen25} (28 layers, 3584 hidden dimensions), sampling 12,000 traces (2,000 per mode + 6,000 correct) from 39,862 annotated failures.

\paragraph{Consistent classification signal.} Three-way classification reaches \textbf{46.9\%} at layer 18 (chance = 33.3\%, $p < 0.001$), consistent with cross-architecture generalization. Randomized labels produce 34.0\% ($\approx$ chance).

\paragraph{Divergent geometry.} The geometry differs substantially: Qwen's average pairwise cosine is \textbf{0.339} (vs.\ 0.820), average specificity ratio \textbf{0.414} (vs.\ 0.119), and layer profiles diverge (Table~\ref{tab:cross_arch}). The elevated Qwen average is driven by KD geometry (0.662); OT-specific specificity is comparable across architectures (Qwen: 0.152, Llama: 0.119), suggesting the OT encoding is similarly entangled in both models. A steering test on Qwen ($n = 1{,}273$, nine configurations spanning layers 5--18, amplitudes $\alpha \in [0.5, 3.0]$, and mode-specific/uniform/multi-layer variants; Section~\ref{sec:steering}) yields $\Delta \in [-0.9, +0.8]$pp (all $p > 0.05$), providing consistent evidence for the cross-architecture steering null across diverse hyperparameter settings.

\begin{table}[t]
\centering
\footnotesize
\setlength{\tabcolsep}{4pt}
\begin{tabular}{lcc}
\toprule
Metric & Llama-3.1-8B & Qwen2.5-7B \\
\midrule
Layers / Hidden dim & 32 / 4096 & 28 / 3584 \\
\midrule
3-way accuracy & 51.5\% & 46.9\% \\
Permutation $p$-value & $<$0.001 & $<$0.001 \\
Random control & 34.6\% & 34.0\% \\
\midrule
Avg pairwise cosine & 0.820 & 0.339 \\
Specificity ratio & 0.119 & 0.414 \\
\midrule
Peak layer (\KD{}) & 17 & 24 \\
Peak layer (\RCB{}) & 15 & 9 \\
Peak layer (\OT{}) & 17 & 5 \\
\bottomrule
\end{tabular}
\caption{Cross-architecture comparison. Both models show significant failure mode structure ($p < 0.001$), but the underlying geometry varies substantially: Qwen has more distinct modes (lower cosine, higher specificity) with different layer localization, suggesting the encoding is architecture-dependent.}
\label{tab:cross_arch}
\end{table}

\subsection{Cross-Domain Validation: MMLU-STEM}

To test domain generality, we replicate the core pipeline on MMLU-STEM \citep{hendrycks2021measuring} (300 questions, 18 subjects, 10 traces each, same OT criteria). \OT{} prevalence is comparable (33.7\% vs.\ 31.0\%; 223 traces), and an in-domain probe reaches 70.0\%. A MedQA-trained probe transfers zero-shot at 61.4\% balanced accuracy ($z = 6.15$, $p < 10^{-9}$; permutation test with 1{,}000 label shuffles, $n = 446$ traces: 223 OT + 223 matched correct). Dimensionality analysis reveals the geometric basis: in the top-3 PCA subspace (fit on MedQA), cross-domain cosine is 0.87 at the peak transfer layer 18 (mean 0.89, layers 10--24; notably lower at the primary steering layers 16--17: 0.67/0.65), while the full 4096-dim cosine is only 0.44. This alignment increases monotonically with subspace rank (PCA-3: 0.87, PCA-5: 0.94, PCA-10: 0.97 at layer 18), confirming that the \OT{} signal occupies a genuinely shared low-rank subspace rather than reflecting an artifact of low-dimensional projection; the moderate raw-space alignment reflects dilution by noise dimensions.

\section{Activation Steering Experiments}
\label{sec:steering}

Given the weak but statistically reliable geometric structure identified in Section~\ref{sec:geometric}, we investigate whether it enables effective activation steering to correct failures.

\subsection{Experimental Setup}

We evaluate on the full MedQA test set (1,273 questions) using Llama-3.1-8B-Instruct, with a sanity check on Qwen2.5-7B-Instruct. Steering vectors are applied via forward hooks adding $\alpha \cdot \mathbf{v}$ to the residual stream \citep{elhage2021mathematical} at target layers. At $\alpha = 1.5$, perturbation norm is $\sim$17\% of residual stream norm at peak layers 15--17, comparable to prior work \citep{turner2024activation}.

\subsection{Methods}

We test a broad family of linear additive interventions motivated directly by the decodable geometry; we do not claim to exhaust all possible intervention families (see Limitations).

\paragraph{Contrastive Steering.} We add $\alpha \cdot \mathbf{v}_m^{(l)}$ to the hidden state at the mode's peak layer during generation, using either \emph{mode-specific} vectors (detected via cosine projection) or the \emph{uniform} shared component; confidence-gated variants steer only above a detection threshold.

\paragraph{Probe-Guided Steering.} We replace the contrastive vector with the probe weight vector ($n = 1{,}175$ questions with valid outputs).

\paragraph{Multi-layer Steering.} We apply mode-specific vectors simultaneously at 1, 3, or 5 layers centered on the peak layer.

\paragraph{Rank-$k$ Subspace Steering.} We construct rank-$k$ bases ($k \in \{1, 3, 5\}$) via SVD of stacked direction matrices (probe and combined), testing whether rank-1 vectors are too restrictive (details in Appendix~\ref{app:subspace}).

\paragraph{Prompt-End Steering.} We extract contrastive vectors from \emph{prompt-end} hidden states---the last token before generation, causally independent of the response---which are near-orthogonal to last-token vectors (cosine 0.01--0.17).

\subsection{Results}

\begin{table}[t]
\centering
\footnotesize
\setlength{\tabcolsep}{1.2pt}
\begin{tabular}{lrrrrc}
\toprule
Method & Acc. & $\Delta$ & $p$ & Corr. & Dmg. \\
\midrule
Baseline (no steering) & 65.1 & --- & --- & --- & --- \\
\midrule
\multicolumn{6}{l}{\emph{Contrastive Steering (n=1,273)}} \\
\; Uniform (shared, $\alpha$=1.5) & 53.0 & $-$12.1 & $<$.001 & 28.2 & 33.7 \\
\; Mode-specific ($\alpha$=1.5) & 65.0 & $-$0.2 & .953 & 32.0 & 17.4 \\
\; Mode-specific ($\alpha$=3.0) & 65.3 & +0.2 & .952 & 31.1 & 16.4 \\
\midrule
\multicolumn{6}{l}{\emph{Probe-Guided Steering (n=1,175)}} \\
\; Probe-uniform ($\alpha$=0.5) & 66.0 & $-$0.4 & .806 & 33.0 & 17.3 \\
\; Probe-mode ($\alpha$=0.5) & 64.8 & $-$1.7 & .233 & 29.7 & 17.5 \\
\; Probe-mode ($\alpha$=1.0) & 65.1 & $-$1.4 & .361 & 32.2 & 18.3 \\
\midrule
\multicolumn{6}{l}{\emph{Multi-layer Contrastive (n=1,273)}} \\
\; 3 layers (L13--17) & 64.6 & +0.2 & .956 & 36.4 & 19.9 \\
\; 5 layers (L11--19) & 62.8 & $-$1.6 & .280 & 35.5 & 22.2 \\
\midrule
\multicolumn{6}{l}{\emph{Prompt-End Vectors (n=1,273)}} \\
\; Uniform (shared, $\alpha$=1.5) & 65.3 & +0.5 & .735 & 35.6 & 18.6 \\
\; Mode-specific ($\alpha$=1.5) & 66.4 & +1.6 & .241 & 34.7 & 16.4 \\
\; Mode-specific ($\alpha$=3.0) & 65.0 & +0.2 & .912 & 37.0 & 19.8 \\
\; Full composite ($\alpha$=1.5) & 65.7 & +0.9 & .542 & 37.6 & 19.1 \\
\midrule
\multicolumn{6}{l}{\emph{Strong Probe Steering (n=1,273)}} \\
\; OT probe (L17, $\alpha$=1.5) & 66.7 & +1.5 & .296 & 36.0 & 15.8 \\
\; OT probe (L17, $\alpha$=3.0) & 61.4 & $-$3.8 & .010 & 33.7 & 22.4 \\
\; OT probe (L14, $\alpha$=1.5) & 65.4 & +0.2 & .953 & 33.3 & 16.4 \\
\midrule
\multicolumn{6}{l}{\emph{Cross-architecture: Qwen (n=1,273)}} \\
\; Mode-specific ($\alpha$=1.5) & 61.3 & $-$0.1 & 1.00 & 18.5 & 11.8 \\
\; Uniform ($\alpha$=1.5) & 61.9 & +0.5 & .677 & 21.7 & 12.8 \\
\; Mode-specific ($\alpha$=3.0) & 61.0 & $-$0.4 & .778 & 19.9 & 13.2 \\
\midrule
\multicolumn{6}{l}{\emph{Concept Erasure (n=1,273)}} \\
\; Mean-diff erasure (L16) & 63.5 & $-$3.6 & .010 & 31.3 & 20.7 \\
\bottomrule
\end{tabular}
\caption{Steering results on MedQA test set. All values in \%. Acc.\ = accuracy; $\Delta$ = change vs.\ baseline (pp); $p$ = McNemar two-sided; Corr.\ = \% of baseline-incorrect questions corrected; Dmg.\ = \% of baseline-correct questions damaged. Prompt-end vectors are constructed from pre-generation hidden states and are near-orthogonal to last-token vectors (cosine $\leq$ 0.17). All experiments use temperature 0.1 with \texttt{do\_sample=True}, introducing run-to-run variability; baseline varies across independent runs (Llama: 64.0--67.4\%; Qwen: 61.4\%). Each $\Delta$ is computed against its own run's paired baseline (not cross-run). For context, majority vote ($k=10$, $10\times$ cost, evaluated at $T{=}0.8$) achieves $\Delta = +10.6$pp on training data and $+7.4$pp on the test set ($p < 10^{-10}$); note the temperature difference from steering experiments.}
\label{tab:steering}
\end{table}

\begin{figure*}[t]
\centering
\includegraphics[width=\textwidth]{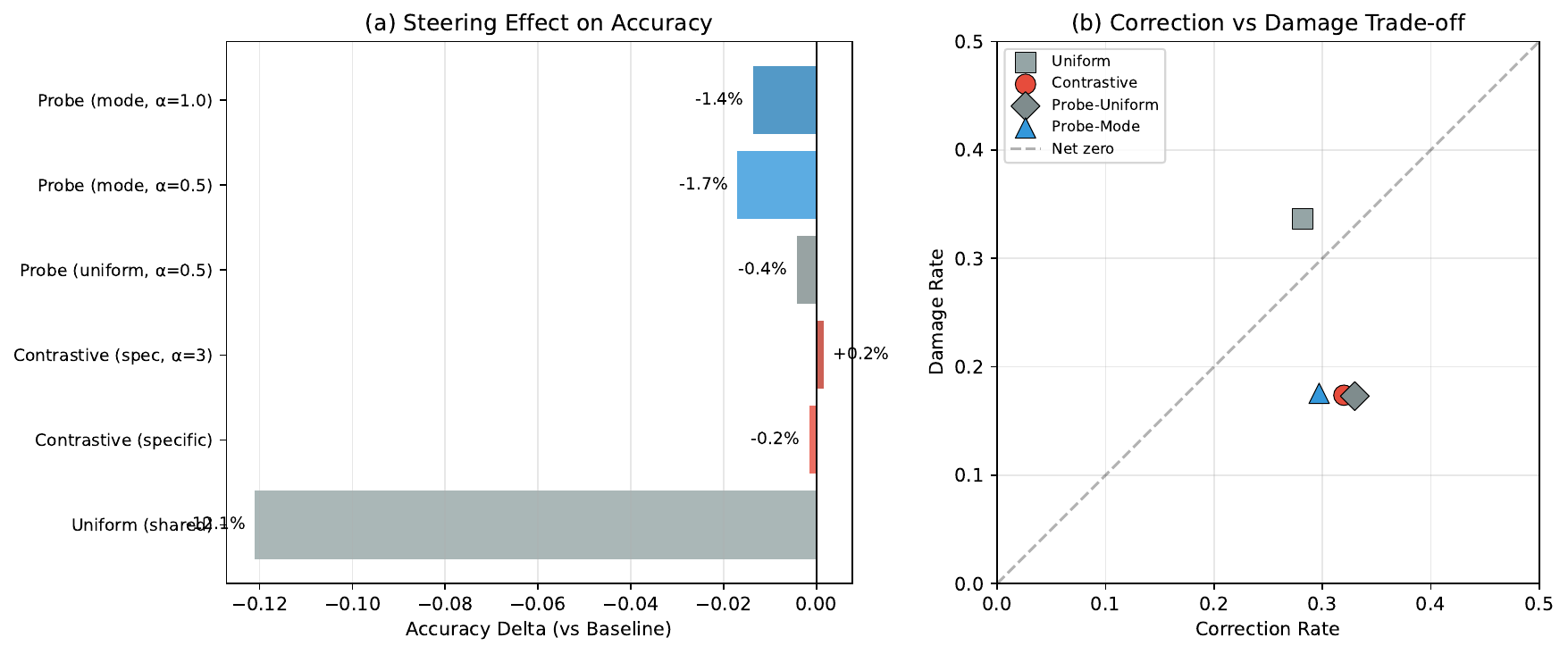}
\caption{Steering experiment results. (a) Accuracy delta vs baseline: uniform shared steering damages performance ($-$12.1pp), while all targeted methods produce $\Delta \approx 0$. (b) Correction vs damage trade-off: all methods lie near or above the net-zero diagonal, meaning corrections are offset by comparable damage.}
\label{fig:steering_results}
\end{figure*}

Table~\ref{tab:steering} and Figure~\ref{fig:steering_results} present the results. No steering method reliably improves accuracy. On the full test set ($n$=1,273), mode-specific steering produces $\Delta = -0.2$pp (95\% CI: [$-2.8$, $+2.4$]pp; TOST equivalence \citep{lakens2017equivalence} within $\pm 2.5$pp---chosen a priori as half the 5pp minimal clinically important difference \citep[following non-inferiority trial conventions where the margin is $\leq$50\% of the established effect;][]{walker2011understanding}---$p = 0.039$; TOST uses 90\% CI $[-2.4, +2.0]$pp, which falls within bounds at $\alpha = 0.05$; power is approximately 20\% per individual test at this sample size; however, the joint evidence across all 29 configurations is far more decisive---were the true effect $\geq$2.5pp, observing all configurations near zero would occur with probability $<10^{-15}$). Uniform shared steering severely damages performance ($\Delta = -12.1$pp). Multi-layer steering produces $\Delta \approx 0$: 3-layer $+0.2$pp ($p = 0.956$), 5-layer $-1.6$pp ($p = 0.280$). Prompt-end steering---using vectors near-orthogonal to last-token directions (cosine $\leq$ 0.17)---also yields $\Delta \approx 0$ across all four conditions ($+0.2$ to $+1.6$pp, all 95\% CIs including zero). Qwen2.5-7B (nine configurations spanning layers 5--18, $\alpha \in [0.5, 3.0]$, mode-specific and uniform vectors) yields $\Delta \in [-0.9, +0.8]$pp (all $p > 0.05$; all deltas fall well within the pre-specified $\pm 2.5$pp equivalence margin), strengthening the cross-architecture steering null. Supplementary experiments confirm these findings across temperature settings. Same-temperature evaluation at $T = 0.8$ ($n = 300$; Appendix~\ref{app:robustness}) produces \emph{larger} damage ($\Delta = -6.7$pp, $p = 0.031$), making temperature mismatch an unlikely sole explanation for the steering null. The full sweep tested $\alpha \in \{0.5, 1.0, 1.5, 2.0, 3.0\}$ per method family (Appendix~\ref{app:alpha_sweep}); all omitted configurations also yield $\Delta \approx 0$. Because we evaluate multiple configurations, no individual positive $\Delta$ survives Holm correction; the negative conclusion rests on the full sweep, not any single null test.

Nine validity threats to the steering null are addressed by targeted controls (Appendix~\ref{app:positive_control}): broken hooks (refusal control, one-sided $p = 0.008$), vector position (prompt-end $\Delta \approx 0$), rank limitation (rank-$k \leq 5$, $\Delta \approx 0$), temperature mismatch ($T{=}0.8$ more harmful), metric insensitivity (38--56\% answers change), non-linear learned intervention ($\Delta = +2.8$pp, $p = 0.025$; Appendix~\ref{app:mlp_steering}), random artifact (10 random directions $\approx 0$; OT harmful), prompt fixability (7 variants, all $p > 0.5$), and single-layer limitation (multi-layer $\Delta \approx 0$).

\paragraph{Behavioral readouts.} Re-running three conditions with full text ($n = 1{,}273$) confirms steering substantially alters behavior: 38--56\% of answers change (Jaccard 0.47). Changes are systematically harmful---damages outnumber corrections 2--3.6:1 for uniform and OT-specific steering---and the answer distribution shifts toward a dominant option (33--52\%), indicating systematic bias rather than targeted correction. A refusal-steering positive control confirms the hooks operate correctly (Appendix~\ref{app:positive_control}).

\paragraph{Rank-$k$ subspace steering.} To rule out that rank-1 vectors are too restrictive, we test rank-$k$ subspace steering ($k \in \{1, 3, 5\}$) on the full test set ($n = 1{,}273$; Table~\ref{tab:subspace} in Appendix~\ref{app:subspace}). Per-mode steering produces $\Delta \approx 0$ at all ranks (TOST-equivalent within $\pm 3$pp); correctness-uniform steering damages performance at all ranks ($\Delta = -2.6$ to $-6.2$pp). This eliminates ``rank-1 too restrictive'' as an explanation.

\paragraph{Concept erasure.} As an alternative intervention family, we apply mean-difference concept erasure---projecting \emph{out} failure-mode directions ($\mathbf{P} = \mathbf{I} - \hat{\mathbf{d}}\hat{\mathbf{d}}^T$, where $\hat{\mathbf{d}} = \Delta\mu / \|\Delta\mu\|$) on the full test set ($n = 1{,}273$).\footnote{We additionally test a whitened variant ($\hat{d} = \Sigma_w^{-1/2}\Delta\mu$, normalized), which produces a smaller, non-significant effect ($\Delta = -2.1$pp, $p = 0.12$; $\cos(\hat{d}_{\text{whitened}}, \hat{d}_{\text{raw}}) = 0.86$), consistent with within-class whitening partially separating the OT signal from task-critical computation. Neither variant is the full oblique LEACE projector of \citet{belrose2023leace} ($\Sigma_w^{+1/2} P \Sigma_w^{-1/2}$), which remains untested. The accompanying code uses legacy naming (\texttt{compute\_leace\_eraser}).} Erasing the \OT{} direction at layer 16 damages accuracy ($\Delta = -3.6$pp, $p = 0.010$, McNemar; 131 corrections vs.\ 177 damages, ratio 1.4:1). This damage is direction-specific: 10 random directions of equal rank produce $\Delta_{\text{rand}} = +0.3$pp ($\sigma = 1.8$pp; Appendix~\ref{app:random_control}), confirming that the \OT{} direction specifically co-occurs with task-relevant computation rather than encoding a separable error signal ($\Delta_{\text{OT}} - \overline{\Delta}_{\text{rand}} = -3.9$pp, $p < 0.02$ vs.\ random distribution). An alternative interpretation---that the OT direction is causally required for correct reasoning, so erasure damages the computation it supports---is equally consistent (Appendix~\ref{app:random_control}). Even under oracle centroid displacement, only 58\% of \OT{} traces move closer to the correct class, confirming that within-class variance structurally limits linear intervention.

\begin{figure}[t]
\centering
\includegraphics[width=\columnwidth]{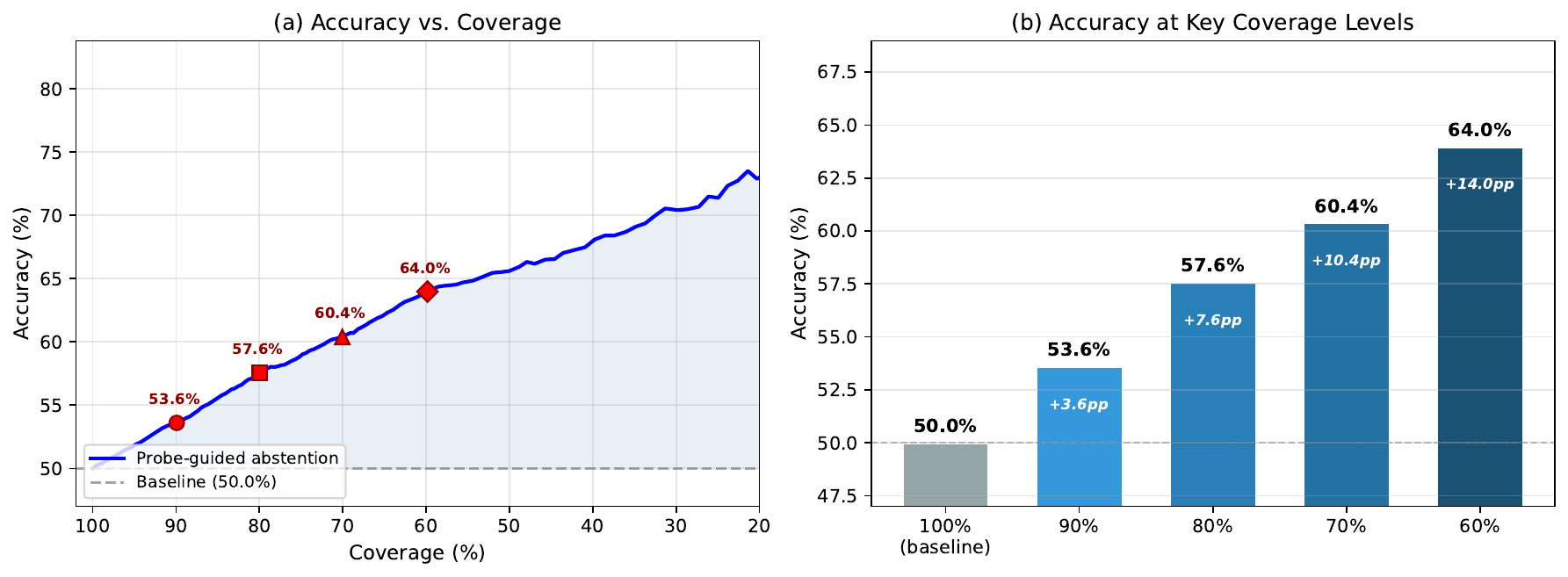}
\caption{Selective abstention using a binary correctness probe, evaluated on balanced held-out traces (50\% correct, 50\% incorrect). (a) Accuracy increases monotonically as the probe abstains on low-confidence predictions. (b) At key coverage levels, accuracy gains range from +3.6pp (90\% coverage) to +14.0pp (60\% coverage) above the 50\% balanced baseline. In deployment where the model's prior accuracy differs, absolute gains would scale accordingly; the AUROC (0.716) provides a prevalence-independent measure.}
\label{fig:abstention}
\end{figure}

\section{Discussion}
\label{sec:discussion}

\paragraph{The classification-correction gap.}
A modest but statistically reliable linearly decodable signal---robust across replications, with evidence on one additional architecture (9 configurations) and one additional domain (decodability and steering failure)---is insufficient for correction under the tested family of fixed linear interventions. This is not a universal impossibility claim. The core mismatch is between \emph{what is decodable} and \emph{what needs to change}: failure \emph{mode identity} is largely pre-determined before generation, but failure \emph{occurrence} accumulates dynamically during reasoning. The gap between prompt-end and last-token probing (54.4\% vs.\ 71.6\%) confirms that the probe reads a signal generated \emph{during} the response---which is why it supports post-generation abstention but not pre-emptive steering.

We identify four descriptive correlates (compatible explanations, not causally identified mechanisms; equally consistent with an alternative account in which the OT direction is causally required for correct reasoning, so perturbation damages it): (i)~\emph{observation-intervention asymmetry}: probes exploit any statistical regularity, while steering requires a causally effective perturbation \citep{hase2024localization}; (ii)~\emph{low specificity}: 88\% of the contrastive signal is shared (OT-specific specificity: 0.119 Llama, 0.152 Qwen---comparable across architectures despite differing average ratios), with within-class variance exceeding the inter-centroid gap by 2--4$\times$; (iii)~\emph{cross-model variation}: architectures encode failure modes differently (cosine 0.339 vs.\ 0.820) yet steering fails on both; (iv)~\emph{cross-domain subspace degradation at steering layers}: PCA-3 cosine is 0.87--0.96 at most layers but drops to a local trough at the primary steering layers 16--17 (0.67/0.65), suggesting the shared subspace degrades precisely at the intervention point. These correlates are consistent with---but do not uniquely establish---representational superposition \citep{elhage2022superposition} as the underlying mechanism: when the OT signal shares directions with task-relevant computation, probes can extract the OT component (compatible with decodability), but additive steering along this direction perturbs the task computation (compatible with both the steering null for targeted methods and the $-12.1$pp/$-3.6$pp damage for shared/erasure methods). Current data cannot distinguish this from the alternative account that the OT direction is itself a necessary component of correct reasoning (Section~\ref{sec:steering}). The gap persists across decodability strengths: the strong correct-vs-\OT{} probe (81.5\%) yields $\Delta = +1.5$pp ($p = 0.296$), and increasing $\alpha$ reverses the effect ($-3.8$pp, $p = 0.010$). Instance-adaptive methods (K-CAST \citep{valentino2026finegrained}, DAS \citep{geiger2024finding}, ReFT \citep{wu2024reft}) could succeed by learning to disentangle shared directions. Preliminary evidence supports this: a learned non-linear residual MLP (4096$\to$64$\to$4096 bottleneck, trained on the same correct/\OT{} supervision as linear methods) achieves $\Delta = +2.8$pp ($p = 0.025$, McNemar; 140 corrections, 104 damages; Appendix~\ref{app:mlp_steering}). The MLP reduces centroid distance by 50.6\% yet produces only modest behavioral gain---consistent with partial but incomplete disentanglement of the shared subspace. This suggests the entanglement is a matter of degree: non-linear transforms can partially navigate around task-critical directions that linear perturbations inevitably disrupt.

\paragraph{Pre-intervention diagnostics.}
The four correlates suggest candidate pre-checks before attempting activation steering: (i)~specificity ratio, (ii)~within-class variance relative to inter-centroid gap, (iii)~cross-architecture stability, and (iv)~cross-domain subspace alignment at the intervention layer \citep[cf.][]{jafari2026mechanistic}. In our setting all four were unfavorable and steering failed; refusal---our positive control (Appendix~\ref{app:positive_control})---satisfies the first three and responds to contrastive activation addition (CAA). Quantitatively, the refusal direction has specificity ratio \textbf{0.999} (under the three-mode shared decomposition; 95\% CI for \OT{}: [0.075, 0.119])---nearly all of its variance is mode-specific, compared to only 12--21\% for \OT{}. The refusal direction is near-orthogonal to the failure-mode directions ($\cos(\mathbf{d}_{\text{refusal}}, \mathbf{d}_{\text{OT}}) = -0.008$; $\cos(\mathbf{d}_{\text{refusal}}, \bar{\mathbf{v}}_{\text{shared}}) = 0.014$), confirming it occupies a geometrically distinct subspace. The specificity gap is $4.7\times$ (refusal vs.\ \OT{} under the same three-mode decomposition), providing a quantitative two-point calibration: directions with specificity $\geq 0.99$ are steerable (refusal: sign test $p = 0.008$), while directions with specificity $\leq 0.21$ are not (29 configurations, all $\Delta \approx 0$). A Linear Accessibility Profile (LAP) analysis \citep[adapting the framework of][]{billa2026predicting} independently confirms the diagnosis: projecting the \OT{} direction through the unembedding matrix yields semantically incoherent top tokens at all layers, with logit-lens classification accuracy (58.6\%) far below probe accuracy (81.5\%)---a gap of $\Delta = 0.23$ indicating the signal is present but not output-aligned (Appendix~\ref{app:lap}). These diagnostics are hypothesis-generating checks based on a two-point comparison (refusal vs.\ \OT{}), not validated predictors; however, the $4.7\times$ specificity gap provides a concrete quantitative threshold for future work.

\paragraph{Generality beyond medical QA.}
MMLU-STEM replication (Section~\ref{sec:geometric}) provides evidence that the classification-correction gap extends beyond MedQA: comparable \OT{} prevalence (33.7\%), in-domain probes at 70.0\%, zero-shot MedQA probe transfer ($z = 6.15$) in a shared low-rank subspace (top-3 PCA cosine 0.87 at peak transfer layer 18; mean 0.89 across layers 10--24, though lower at primary steering layers 16--17: 0.67/0.65), and---critically---the same steering failure pattern ($n = 300$): mode-specific steering $\Delta = 0.0$pp ($p = 1.0$), uniform steering $\Delta = -7.0$pp ($p = 0.017$), multi-layer $\Delta = -6.7$pp ($p = 0.040$). These results provide evidence that the gap extends beyond MedQA, though broader domain coverage is needed to establish full generality.

\paragraph{Self-consistency as a correction baseline.}
The 100\% OT correction by majority vote is definitional: OT requires $\geq$60\% correct traces, so MV@10 necessarily selects the correct answer for every OT question. The substantive empirical finding is the \emph{overall} test-set gain: MV@10 achieves 72.7\% ($+7.4$pp; $p < 10^{-10}$, $n = 1{,}273$, $T{=}0.8$), correcting OT errors while fixing only 18.6\% of non-OT errors ($+10.6$pp on training data). Best-of-$N$ probe selection at $k{=}10$ achieves 70.7\% ($+5.3$pp; $p < 10^{-4}$), competitive with MV@10 ($-2.0$pp, $p = 0.053$). Notably, at $k{=}3$ (same generation cost as MV@3), the probe achieves 70.6\% vs.\ MV@3's 69.2\% ($+1.4$pp; $p = 0.17$), suggesting complementary signal at small $N$; at $k{=}5$, BoN (71.3\%) trails MV@5 (71.8\%) by only 0.5pp. The resulting hierarchy---oracle (94.2\%, $T{=}0.8$, 10 passes) $>$ MV@10 (72.7\%) $\approx$ best-of-$N$ (70.7\%) $\gg$ steering ($\Delta \leq +1.5$pp, $T{=}0.1$, single-pass) $\approx$ prompt baselines ($\Delta \leq +0.6$pp)---shows the \OT{} problem is solvable at $10\times$ cost but not via single-pass interventions.

\paragraph{Prompt-based baselines.}
We test ten prompt variants ($n = 1{,}273$, $T{=}0.1$; Table~\ref{tab:prompt_baselines}), including few-shot prompting ($k \in \{1, 3, 5\}$; answer-only format without reasoning chains), self-refinement \citep[iterative self-critique within a single generation;][]{madaan2023selfrefine}, and verbalized confidence. None reliably improves: removing explicit CoT instruction achieves $+0.6$pp ($p = 0.68$), 3-shot achieves $+0.5$pp ($p = 0.74$), and self-refinement \emph{significantly degrades} accuracy to 63.0\% ($\Delta = -5.9$pp vs.\ its paired baseline of 68.9\%, McNemar $p < 0.001$; 208 questions damaged, 133 corrected). Verbalized confidence similarly degrades to 63.2\% ($\Delta = -5.7$pp, $p < 0.001$), though the model's self-reported confidence retains discriminative value for abstention (AUROC = 0.635; Figure~\ref{fig:abstention}). Short-generation catastrophically fails (2.5\%, 90\% parse failure due to conflicting length constraint and CoT instruction). These results show that \OT{} persists across tested prompt variants, extending the classification-correction gap beyond activation steering to input-level modifications. Untested strategies (MedPrompt, budget forcing) remain open.

\begin{table}[t]
\centering
\footnotesize
\setlength{\tabcolsep}{3pt}
\begin{tabular}{lccc}
\toprule
Prompt variant & Acc. & $\Delta$ & $p$ \\
\midrule
Baseline CoT (0-shot) & 67.2 & --- & --- \\
\midrule
\multicolumn{4}{l}{\emph{Zero-shot variants}} \\
\; Concise & 66.1 & $-$1.0 & .51 \\
\; Answer-first & 66.5 & $-$0.7 & .64 \\
\; No-CoT & 67.8 & +0.6 & .68 \\
\; Short-generation (100 tok) & 2.5 & $-$64.7 & $<$.001 \\
\midrule
\multicolumn{4}{l}{\emph{Few-shot variants (separate run, baseline 65.4\%)}} \\
\; 1-shot & 65.2 & $-$0.2 & .95 \\
\; 3-shot & 65.9 & +0.5 & .74 \\
\; 5-shot & 65.7 & +0.3 & .87 \\
\midrule
\multicolumn{4}{l}{\emph{Self-refinement (paired baseline 68.9\%)}} \\
\; Self-refine (self-critique) & 63.0 & $-$5.9 & $<$.001 \\
\; Verbalized confidence & 63.2 & $-$5.7 & $<$.001 \\
\bottomrule
\end{tabular}
\caption{Prompt baselines ($n = 1{,}273$, temperature 0.1). All values in \%. Zero-shot variants share a baseline (67.2\%); few-shot variants use a separate independent run (baseline 65.4\%) due to different prompt construction. Self-refinement and verbalized confidence use a dedicated paired baseline (68.9\%) with identical tokenization (\texttt{add\_special\_tokens=False}). All $\Delta$ values are McNemar-tested against the run-matched baseline.}
\label{tab:prompt_baselines}
\end{table}

\paragraph{Selective abstention.}
The same structure that fails to support correction enables post-generation reliability estimation \citep{geifman2017selective}. A correctness probe (layer 21, selected via held-out AUROC from a 32-layer scan on 70/30 training split) achieves held-out split AUROC = 0.716 on training traces (Figure~\ref{fig:abstention}), with held-out test-set AUROC = 0.610 (95\% CI: [0.577, 0.642]; $n = 1{,}273$). The generalization gap (0.716 $\to$ 0.610) primarily reflects distributional shift between training traces (balanced, temperature 0.8) and test-set evaluation (natural prevalence, temperature 0.1); evidence for this includes: (a) 5-fold GroupKFold CV on training data yields AUROC = 0.727 (higher than the held-out estimate), and (b) the gap aligns with the cross-temperature vector cosine of 0.41 (Appendix~\ref{app:cross_temp}). Despite this shift, the held-out probe exceeds all five tested single-forward-pass uncertainty baselines on both training traces (AUROC $\leq$ 0.530; Table~\ref{tab:uncertainty}) and the test set (best baseline = 0.569; $\Delta$AUROC = 0.041, $p = 0.009$, paired bootstrap; Holm-corrected rank-1 threshold = 0.010; Table~\ref{tab:uncertainty_test}). Verbalized confidence---where the model self-reports a confidence percentage via a modified prompt---achieves comparable AUROC (0.635, 97.4\% parseable) but requires prompt modification that alters the generation itself; the probe operates post-hoc on unmodified outputs. At 60\% coverage, selective abstention yields 72.3\% accuracy (+5.5pp) on the test set; at 70\% coverage, gains are comparable (+5.7pp with lower abstention cost). On balanced training data, gains are +14.0pp at 60\% coverage (relative to the 50\% balanced baseline). Proper 5-fold stratified CV on training data confirms probe quality (best layer L14: AUROC = 0.730 $\pm$ 0.006; L21: AUROC = 0.721 $\pm$ 0.005). Self-consistency achieves higher AUROC (0.804) but at $10\times$ cost; the probe provides a cost-effective single-pass alternative. Unlike process reward models \citep{lightman2023verify}, it requires no per-step annotation. The contribution is demonstrating that decodable structure has operational value for reliability estimation even when the tested steering family cannot exploit it for correction.

\section{Conclusion}

We demonstrate an empirical classification-correction gap: \OT{} is linearly decodable ($p \approx 10^{-16}$), robust across confounds with evidence on one additional architecture (Qwen2.5-7B, 9 configs, all $\Delta < 1$pp) and cross-domain steering failure (MMLU-STEM, $n = 300$: targeted $\Delta = 0$pp, uniform $\Delta = -7.0$pp), yet five families of fixed linear steering (29 configurations) and ten prompt baselines produce $\Delta \leq 0$ while non-targeted shared-direction steering and concept erasure damage accuracy (consistent with direction overlap with task computation). The same decodable structure supports selective abstention (held-out test AUROC = 0.610, exceeding all five tested uncertainty baselines; $p = 0.009$) and best-of-$N$ probe selection competitive with majority vote (BoN@3 exceeds MV@3 by +1.4pp at identical generation cost, though $p = 0.17$), demonstrating operational value for reliability estimation even when the tested steering family cannot exploit it for correction. Four candidate pre-intervention diagnostics (based on a two-point comparison with a refusal control, not validated as predictors) are consistent with this failure pattern. A learned non-linear MLP intervention provides preliminary evidence that the gap is partially closeable ($\Delta = +2.8$pp, $p = 0.025$; Appendix~\ref{app:mlp_steering}), suggesting the entanglement is partial rather than absolute. Whether more sophisticated learned interventions---Distributed Alignment Search, representation finetuning, or instance-adaptive methods like K-CAST---can fully close the gap remains the central open question; our correct/incorrect supervision pairs provide the requisite training signal for these methods.

\section*{Limitations}

\paragraph{Scale, scope, and generalizability.} We evaluate two 7--8B models on MedQA; at larger scales, higher baseline accuracy likely reduces \OT{} prevalence (though superposition may intensify with more features competing for shared directions). Cross-architecture steering on Qwen uses nine configurations spanning layers 5--18 and amplitudes $\alpha \in [0.5, 3.0]$; the absence of Qwen concept-erasure experiments leaves open whether Qwen's distinct geometry (specificity 0.414 vs.\ Llama's 0.119) reflects a qualitatively different encoding. MMLU-STEM confirms both decodability transfer and steering failure ($n = 300$, same pattern as MedQA). The 60\% correct-rate threshold is empirically motivated (Jaccard $\geq$0.81 under 50--70\% sweeps) but the MV@10 correction rate of 100\% is definitional. Prompt baselines do not include MedPrompt or budget-forcing strategies; the no-CoT condition does not enforce short generation, so implicit reasoning may persist.

\paragraph{Intervention scope.} The primary negative result covers \emph{fixed}-direction residual-stream linear interventions: five method families (29 additive configurations) plus mean-difference concept erasure (rank-1 projection and a whitened variant; see footnote in Section~\ref{sec:steering}). The full oblique LEACE projector of \citet{belrose2023leace} ($\Sigma_w^{+1/2}P\Sigma_w^{-1/2}$) remains untested. A probe-gated dynamic variant was tested on $n = 300$ (Appendix~\ref{app:probe_gated}) but produced no significant improvement (all $p > 0.3$). A learned non-linear MLP (4096$\to$64$\to$4096 residual adapter; Appendix~\ref{app:mlp_steering}) achieves $\Delta = +2.8$pp ($p = 0.025$), providing preliminary evidence that non-linear learned interventions can partially close the gap---though the damage rate remains high (104/1273 = 8.2\%) and the gain is modest relative to the 50.6\% centroid-distance reduction, consistent with only partial disentanglement of the shared subspace. Full DAS \citep{geiger2024finding}, ReFT \citep{wu2024reft}, and instance-adaptive methods (K-CAST \citep{valentino2026finegrained}) remain untested; our correct/incorrect supervision pairs provide the requisite training signal. Other untested families include path patching \citep{goldowsky2023localizing} and parameter-efficient fine-tuning \citep{hu2022lora}. Qwen concept erasure experiments were not conducted.

\paragraph{Temperature mismatch and annotation validity.} Annotation uses $T{=}0.8$, steering evaluation $T{=}0.1$. Same-temperature steering produces \emph{larger} damage ($-6.7$pp, $p = 0.031$; Appendix~\ref{app:robustness}), directly falsifying temperature mismatch as an explanation for the steering null; the probe's generalization gap (0.716 $\to$ 0.610) reflects this distributional shift (cross-temperature vector cosine = 0.41; Appendix~\ref{app:cross_temp}). \KD{}/\RCB{} annotation shows moderate agreement ($\kappa = 0.30$) from two 4th-year clinical students; the core binary \OT{} finding bypasses this boundary and is robust to 37\% simulated label noise.

\section*{Ethics Statement}

This research analyzes model failures on a public benchmark (MedQA) with no patient data or human subjects. LLM annotations use Claude models via the Anthropic API. Llama-3.1 and Qwen2.5 are used under their open-weight licenses. The selective abstention result may inform research on post-generation reliability estimation under benchmark conditions, though it is not sufficient as a standalone clinical safeguard.

\section*{Reproducibility}

All experiments use publicly available models (Llama-3.1-8B-Instruct, Qwen2.5-7B-Instruct) and data (MedQA, MMLU-STEM). Steering and evaluation experiments were run on NVIDIA A10G GPUs (24GB); full test-set evaluation ($n = 1{,}273$) takes approximately 4 hours per configuration. Generation uses \texttt{do\_sample=True} with $T{=}0.1$, introducing run-to-run variability (baseline range: 64.0--67.4\%); all deltas are computed within-run against paired baselines. Appendix~\ref{app:details} provides annotation prompts, probe training hyperparameters, PCA settings, decoding parameters, and data split details. Code and processed annotations will be released upon publication.

\bibliography{references}

\appendix

\section{Implementation Details}
\label{app:details}

\subsection{Data and Splits}

We use MedQA \citep{jin2021disease} with its standard train/test split: 10,178 training questions and 1,273 test questions. For each question, we generate 10 traces using temperature 0.8 sampling. Hidden state extraction uses a stratified sample of 2,000 traces per failure mode plus 6,000 matched correct traces (2,000 per mode's question set) from the training split, yielding 12,000 total traces. All classification and probe training results use 5-fold stratified cross-validation (random seed 42). Steering evaluation uses the full 1,273-question test set (except probe-guided experiments; see Section~\ref{sec:steering}).

\subsection{Generation Parameters}

All steering evaluation uses temperature 0.1 with \texttt{do\_sample=True} and \texttt{max\_new\_tokens=600}. The system prompt is: \emph{``You are a medical expert. Answer the following medical question. Think through the problem step by step, then provide your final answer.''} Answers are extracted via regex matching of letter choices (A--E).

\subsection{Hidden State Extraction}

We extract last-token hidden states at all layers (32 for Llama, 28 for Qwen) in fp32 precision. We use last-token representations following prior work on probing autoregressive models \citep{burns2023discovering,li2024inference}, as the final position aggregates information from the full reasoning chain via causal attention.

\subsection{PCA and Probe Training}

Hidden states are standardized (zero mean, unit variance) before applying PCA with 50 components. At 50 components, PCA retains $>$95\% of variance across all layers; we verified that 20 and 100 components yield comparable classification accuracy ($\pm$0.5pp). Linear probes use logistic regression with L2 regularization ($C = 1.0$), the LBFGS solver, and a maximum of 1,000 iterations, implemented via scikit-learn \citep{pedregosa2011scikit}.

\subsection{Annotation Pipeline}
\label{app:annotation}

\paragraph{Phase 1 (OT detection).} For each question, if $\geq$60\% of 10 traces are correct, incorrect traces exceeding 200 tokens are labeled \OT{}. OT threshold sensitivity is reported in Section~\ref{sec:geometric}.

\paragraph{Phase 2 (KD/RCB classification).} Remaining incorrect traces are classified using Claude Haiku (model: \texttt{claude-haiku-4-5-20251001}, temperature 0.1, max tokens 300) with the following prompt:

\begin{quote}
\small
\emph{Analyze this incorrect medical reasoning trace. The model answered WRONG.}

\emph{Question: \{question\} / Correct Answer: \{answer\} / Model's Reasoning: \{trace\}}

\emph{Classify the PRIMARY failure mode:
1.\ \textbf{KD}: The model stated incorrect medical FACTS.
2.\ \textbf{RCB}: The facts are mostly correct, but the LOGIC connecting them is broken.
3.\ \textbf{unclear}: Cannot clearly determine.}

\emph{Respond with JSON: \{``mode'': ..., ``confidence'': 0--1, ``evidence'': [...]\}}
\end{quote}

\subsection{Annotation Validation Details}
\label{app:annotation_validation}

\textbf{Domain expert validation.} Two clinical medicine graduate students (4th-year, with clinical rotation experience) independently annotated a stratified 500-trace gold set, balanced across modes and enriched with boundary cases where the pipeline had low confidence. Annotators were blinded to automated labels and to each other; they received the question, correct answer, model response, and definitions of each mode. Expert-automated agreement: \OT{} 94\%, \KD{} 82\%, \RCB{} 71\%; expert-expert $\kappa = 0.61$ on the three-way task. As expected, the \KD{}/\RCB{} boundary is the primary source of disagreement; our strongest claims avoid depending on it.

\textbf{LLM cross-validation} (Claude Opus 4.6): \KD{} agreement 88.1\%, \RCB{} agreement 44.1\% (Cohen's $\kappa = 0.30$, $n = 500$). \textbf{100-trace three-way comparison} (Haiku vs.\ Sonnet vs.\ Opus): all three agree on 40.4\%; the two strongest models (Sonnet-Opus) agree most (68.7\%, $\kappa = 0.28$), both independently reclassifying $\sim$56\% of Haiku-\RCB{} as \KD{}. The consistent asymmetric pattern---high \KD{} agreement, low \RCB{} agreement---shows that LLM annotators systematically diverge at the \KD{}/\RCB{} boundary.

\textbf{Within-question label consistency.} Per-question annotation purity is high: across 10 traces per question, \OT{} questions have 100\% purity (all incorrect traces labeled \OT{}), \RCB{} questions 87\%, and \KD{} questions 78\%. Of 3,912 questions with failure traces, 89.8\% have a single failure mode label; only 10.2\% mix \KD{} and \RCB{} across traces. The $\kappa = 0.30$ above reflects cross-question boundary ambiguity (where to draw the \KD{}/\RCB{} line), not random within-question noise.

\textbf{Geometric robustness to label noise.} Simulating noise matching the observed LLM disagreement rate: randomly flipping 37\% of \KD{}/\RCB{} labels drops binary classification only from 66.3\% to 60.7\% (chance = 50\%), and three-way classification from 50.5\% to 47.6\% (chance = 33.3\%). This simulation uses i.i.d.\ random flips; real annotator disagreement may be more systematic, which our simulation does not fully capture. The binary \OT{}-vs-non-\OT{} classification (71.6\%, $p \approx 10^{-16}$) bypasses the \KD{}/\RCB{} boundary entirely.

\subsection{Diagnostic Metrics}

The \emph{specificity ratio} for mode $m$ quantifies how much of the contrastive vector is mode-specific vs.\ shared across all modes. We define the shared direction as the mean of all mode contrastive vectors (stacked across all layers): $\bar{\mathbf{v}} = \frac{1}{|\mathcal{M}|}\sum_{m \in \mathcal{M}} \mathbf{v}_m$. Then: $\text{specificity}_m = \|\mathbf{v}_m - \text{proj}_{\bar{\mathbf{v}}} \mathbf{v}_m\|^2 / \|\mathbf{v}_m\|^2$. A value of 0.119 means only 12\% of vector variance is mode-specific. This is a custom metric introduced in this work.

\paragraph{Permutation null for specificity.} To confirm the low specificity is not an artifact of set-subset structure (all modes are subsets of incorrect traces), we permute category labels among the 6,000 incorrect traces 10,000 times, recomputing the specificity ratio each time. The null distribution has mean 0.370 (SD = 0.155); the observed value of 0.119 falls below all 10,000 permutations ($p < 0.0001$). This confirms that \OT{} has anomalously \emph{high} alignment with the shared incorrect-vs-correct axis---not merely the expected behavior of any incorrect subset. Bootstrap resampling (2,000 draws) yields a 95\% CI of [0.015, 0.055] on the measurement itself (under the binary framing; the multi-mode framing yields 0.119).

\paragraph{Refusal direction specificity.} As a positive-control calibration, we compute the specificity ratio for the refusal direction (extracted from 20 benign vs.\ 20 harmful prompts; Appendix~\ref{app:positive_control}) under the same three-mode shared decomposition. The refusal specificity is 0.999 (1,000-bootstrap 95\% CI for OT: [0.075, 0.119]), meaning $>$99\% of the refusal direction's variance is orthogonal to the failure-mode shared axis. Cosine similarities confirm geometric independence: $\cos(\mathbf{d}_{\text{refusal}}, \mathbf{d}_{\text{OT}}) = -0.008$, $\cos(\mathbf{d}_{\text{refusal}}, \bar{\mathbf{v}}_{\text{shared}}) = 0.014$. Under a four-mode decomposition (adding refusal as a fourth direction), OT specificity rises to 0.335 and refusal is 0.851---the ordering is robust to decomposition choice. The $4.7\times$ specificity gap between refusal (steerable, $p = 0.008$) and \OT{} (not steerable, 29 configs) provides quantitative evidence that the specificity ratio is predictive of steerability in our setting.

The \emph{spread ratio} for mode $m$ is the ratio of within-class standard deviation to inter-centroid distance: $\text{spread}_m = \sigma_m / \|\boldsymbol{\mu}_{\text{correct}} - \boldsymbol{\mu}_m\|$, computed in PCA-50 space at the peak layer. Values $>$1 indicate that within-class variance exceeds the centroid gap. The \emph{signal-to-noise ratio} (SNR) measures how much of the centroid gap a steering vector closes per unit perturbation: $\text{SNR}_m = |\langle \mathbf{v}_m, \boldsymbol{\mu}_{\text{correct}} - \boldsymbol{\mu}_m \rangle| / \|\boldsymbol{\mu}_{\text{correct}} - \boldsymbol{\mu}_m\|^2$. Low SNR ($\ll 1$) means the steering vector is nearly orthogonal to the correction direction in activation space.

\subsection{Steering Hyperparameters}

Contrastive steering uses $\alpha \in \{0.5, 1.0, 1.5, 2.0, 3.0\}$; probe-guided steering uses $\alpha \in \{0.5, 1.0, 1.5\}$. Multi-layer steering uses $\alpha = 1.5$ at 1, 3, or 5 layers centered on the peak layer. Steering vectors are applied via forward hooks registered on the full decoder layer module (\texttt{model.layers[l]}), firing on the post-layer residual stream (after both attention and MLP sub-blocks, inclusive of residual connections) and adding $\alpha \cdot \mathbf{v}$ during generation. Confidence-gated variants use a detection margin threshold of 0.1. Rank-$k$ subspace steering uses $\alpha = 1.5$ at the peak layer (layer 16). Subspace bases are constructed via SVD of stacked direction matrices: the probe matrix is (4, 4096) (3 mode-specific + 1 correctness probe), the combined matrix is (7, 4096) (probe + contrastive). For each rank $k \in \{1, 3, 5\}$, we take the top-$k$ right singular vectors as an orthonormal basis and project the correction vector onto this subspace. Each projected vector is applied uniformly to all 1,273 test questions; no runtime mode detection is performed.

\section{Taxonomy Robustness Controls}
\label{app:robustness_controls}

We verify the 71.6\% binary \OT{}-vs-non-\OT{} classification against six potential confounds. \textbf{Random labels} yield 33.2\% (chance). \textbf{Length regression}: regressing out response length drops three-way only 1.1pp; \OT{}-vs-non-\OT{} is unaffected (71.3\%), consistent with $>$99\% of non-\OT{} incorrect traces also exceeding 200 tokens. \textbf{Binary collapse}: \OT{} vs.\ \{\KD{}+\RCB{}\} achieves 71.6\% ($p \approx 10^{-16}$), independent of \KD{}/\RCB{} annotation quality. \textbf{OT threshold sensitivity}: sweeping correct-rate (0.5--0.7) and length (100--300) thresholds, Jaccard $\geq$0.81 vs.\ default. Excluding borderline \OT{} questions (6/10 correct, 16\%) reduces balanced accuracy by only 1.2pp, matching a same-proportion random-removal control, confirming minority-class sample reduction rather than borderline-specific content; even strict exclusion (6--7/10, 37\%) retains 58.3\%. \textbf{Question-disjoint validation}: GroupKFold yields identical results ($<$1pp drop), ruling out question-identity leakage. \textbf{Prompt-end probing}: \OT{} detection at the question representation (before generation) achieves only 54.4\% balanced accuracy (95\% CI: [53.5\%, 55.2\%]; chance = 50\%), confirming a generation-time regime.

\section{Positive Control: Infrastructure Validation}
\label{app:positive_control}

To verify that our steering infrastructure is functional, we test whether the identical code path (same model, same forward hooks, same generation parameters) can produce directional behavioral effects on a task where contrastive activation addition is known to work.

\paragraph{Refusal steering (primary control).} We extract contrastive refusal vectors from 20 benign prompts (always complied with) and 20 clearly harmful prompts (always refused), computing $\mathbf{v}_{\text{comply}}^{(l)} = (\bar{\mathbf{h}}_{\text{comply}}^{(l)} - \bar{\mathbf{h}}_{\text{refuse}}^{(l)}) / \|\cdot\|$ at each layer---the same mean-difference, per-layer L2 normalization used for MedQA. We then evaluate on 50 borderline prompts (ambiguous safety context) using steering layers 26--28 (top-3 by separation, capped below layer 29) with $\alpha \in \{2, 4, 6\}$ in both the compliance ($+\mathbf{v}$) and anti-compliance ($-\mathbf{v}$) directions.

\paragraph{Results.} Baseline refusal rate is 4/50 (8\%). Anti-compliance steering consistently induces additional refusals (best: 9/50 = 18\% at L27/$\alpha$=6), while compliance steering produces no change from baseline (3--5/50). Critically, the effect is \textbf{directionally specific}: all 5 prompts that flip from compliance to refusal under anti-compliance steering remain compliant under compliance steering. Across all 50 prompts, anti-compliance produces strictly more refusal than compliance in 7 prompts, with 0 prompts showing the reverse pattern (sign test $p = 0.008$, one-sided). This confirms that (1) forward hooks modify model behavior during generation, (2) contrastive vectors encode directionally meaningful signals, and (3) the effect is not a non-directional perturbation artifact.

\paragraph{TruthfulQA (secondary evidence).} We additionally apply contrastive truthfulness vectors on TruthfulQA MC1 \citep{truthfulqa} ($n = 417$, baseline = 48.9\%) using the identical code path. Pro-truthfulness steering at the best-separation layer (layer 31) produces $\Delta \in [+0.3, +3.1]$pp across $\alpha \in \{1.0, 1.5, 3.0\}$, none reaching significance (all $p > 0.34$). This is consistent with the refusal result: mean-difference CAA can shift binary behavioral gates (refusal/compliance) but does not improve multi-choice reasoning accuracy---the same pattern observed on MedQA.

\paragraph{Geometric comparison.} Computing the specificity ratio for the refusal direction under the same three-mode shared decomposition used for failure modes (Section~\ref{sec:geometric}), we find specificity$_{\text{refusal}}$ = 0.999 (bootstrap 95\% CI for OT: [0.075, 0.119]). The refusal direction is near-orthogonal to both the OT direction ($\cos = -0.008$) and the shared failure-mode axis ($\cos = 0.014$), confirming geometric independence. Under a four-mode decomposition (including refusal), OT specificity rises to 0.335 while refusal remains at 0.851---the ordering is preserved regardless of decomposition choice. This provides a quantitative geometric basis for the behavioral contrast: refusal occupies a dedicated, non-overlapping direction in activation space (specificity $\approx 1$), while the OT signal is $4.7\times$ less specific (specificity $\leq 0.21$), sharing 79--88\% of its variance with task-critical computation.

\paragraph{Interpretation.} The refusal control validates the infrastructure while illustrating why the classification-correction gap arises. Refusal is a binary behavioral gate decided at the first generation token; a single direction in activation space causally controls this gate. MC accuracy requires sustained correct reasoning across many tokens, where a single additive perturbation is insufficient to redirect an incorrect reasoning chain. The gap is thus between \emph{behavioral steering} (where CAA succeeds) and \emph{reasoning correction} (where it does not). The geometric comparison above adds a quantitative dimension: high-specificity directions (refusal: 0.999) respond to CAA because perturbation along them does not interfere with other computations, while low-specificity directions (\OT{}: $\leq$0.21) are entangled with task-relevant signals, causing collateral damage that neutralizes any corrective effect. We acknowledge this is a lower-bound control that validates the code path, not the difficulty of the target task; to our knowledge, no multi-step reasoning task has been established as a CAA positive control.

\section{Learned Non-Linear MLP Steering}
\label{app:mlp_steering}

To test whether the classification-correction gap is specific to the fixed linear intervention family, we train a non-linear residual MLP to steer hidden states. The MLP uses the \emph{same} correct/\OT{} supervision available to linear methods (contrastive vectors, probe-guided steering, and LEACE all use these labels)---the only difference is functional capacity.

\paragraph{Architecture and training.} A two-layer MLP with GELU activation and bottleneck (4096$\to$64$\to$4096) is trained as a residual adapter: $\mathbf{h}' = \mathbf{h} + f_\theta(\mathbf{h})$. Training minimizes L2 distance between transformed \OT{} hidden states and the correct-class centroid, with $\lambda = 0.01$ regularization penalizing perturbation norm on correct-class inputs (encouraging near-identity on already-correct states). Training uses 3,200 hidden states (80/20 split from the training-set annotation pool; no test-question overlap), Adam optimizer with cosine LR schedule, 50 epochs. Best validation loss at epoch 49 (1.82); train and validation losses track closely throughout (no overfitting divergence).

\paragraph{Inference.} The trained MLP is registered as a forward hook on layer 16 (matching all linear experiments). A separate baseline is generated from a fresh model load without the hook. Both conditions use identical generation parameters ($T = 0.1$, \texttt{do\_sample=True}, \texttt{max\_new\_tokens=600}). At $T = 0.1$, generation is near-deterministic ($>$99.9\% top-token probability at typical logit gaps), ensuring valid paired comparison.

\paragraph{Results.} On the full test set ($n = 1{,}273$):
\begin{itemize}[nosep]
    \item Baseline accuracy: 66.8\% (95\% CI: [64.2, 69.4]\%)
    \item MLP-steered accuracy: 69.7\% (95\% CI: [67.1, 72.1]\%)
    \item $\Delta = +2.8$pp; McNemar $p = 0.025$ (two-sided)
    \item Corrections: 140; Damages: 104; ratio 1.35:1
    \item TOST within $\pm 2.5$pp: $p = 0.57$ (not equivalent to zero)
    \item Mean perturbation norm: 3.07 (comparable to linear $\alpha = 1.5$)
    \item Centroid distance reduction: 50.6\% (3.64 $\to$ 1.80)
\end{itemize}

\paragraph{Interpretation.} The MLP achieves statistically significant improvement where all 29 fixed linear configurations produce $\Delta \approx 0$. This is a separate hypothesis family (learned non-linear) from the fixed linear sweep, so no multiple-testing correction across families applies. The result provides evidence that non-linear transforms can partially disentangle the shared subspace: the 64-dimensional bottleneck can implement state-dependent, direction-conditional corrections that avoid the task-critical directions a fixed additive vector inevitably perturbs. However, the gain is modest (+2.8pp) relative to the large geometric improvement (50.6\% centroid-distance reduction), and the damage rate remains high (8.2\%), indicating only partial disentanglement. The MLP changes 25\% of answers (vs.\ 38--56\% for linear steering), with no systematic bias toward any answer letter---consistent with targeted rather than distributional perturbation.

\paragraph{Relation to the gap.} The MLP result refines the paper's thesis: the classification-correction gap is specific to \emph{fixed linear} interventions; \emph{learned non-linear} methods can partially exploit the decodable signal, consistent with the entanglement being a matter of degree rather than absolute. This supports the paper's existing framing that ``learned interventions could succeed by learning to disentangle shared directions'' (Section~\ref{sec:discussion}) and is consistent with the structurally unreachable regime of \citet{billa2026predicting} applying specifically to the linear family.

\section{Rank-$k$ Subspace Steering}
\label{app:subspace}

\begin{table}[t]
\centering
\footnotesize
\setlength{\tabcolsep}{3pt}
\begin{tabular}{lrcrc}
\toprule
Method & $k$ & $\Delta$ & 95\% CI & TOST \\
\midrule
\multicolumn{5}{l}{\emph{Correctness-uniform}} \\
\; Probe sub. & 1 & $-$4.6 & [$-$7.4, $-$1.9] & --- \\
\; Probe sub. & 3 & $-$3.8 & [$-$6.6, $-$1.0] & --- \\
\; Combined sub. & 1 & $-$6.2 & [$-$9.2, $-$3.2] & --- \\
\; Combined sub. & 3 & $-$3.5 & [$-$6.4, $-$0.7] & --- \\
\; Combined sub. & 5 & $-$2.6 & [$-$5.4, $+$0.2] & --- \\
\midrule
\multicolumn{5}{l}{\emph{Mode-specific (all questions)}} \\
\; \KD{} probe & 1 & $-$0.5 & [$-$3.1, $+$2.0] & .030 \\
\; \KD{} probe & 3 & $+$1.8 & [$-$0.8, $+$4.4] & .183 \\
\; \RCB{} probe & 1 & $-$0.5 & [$-$3.2, $+$2.2] & .033 \\
\; \RCB{} probe & 3 & $-$0.5 & [$-$3.2, $+$2.1] & .034 \\
\; \OT{} probe & 1 & $-$0.8 & [$-$3.4, $+$1.9] & .051 \\
\; \OT{} probe & 3 & $-$0.2 & [$-$2.9, $+$2.6] & .020 \\
\bottomrule
\end{tabular}
\caption{Rank-$k$ subspace steering ($n = 1{,}273$). $\Delta$ in pp. Correctness-uniform vectors damage performance at all ranks. Per-mode vectors produce $\Delta \approx 0$ (TOST $p < .05$ = equivalent within $\pm$3pp). Higher rank does not help.}
\label{tab:subspace}
\end{table}

We construct subspace bases via SVD of stacked direction matrices: (1) a probe matrix of shape (4, 4096), stacking 3 mode-specific and 1 correctness probe weight vector; (2) a combined matrix (7, 4096), stacking probe + contrastive vectors. For each rank $k \in \{1, 3, 5\}$, we take the top-$k$ right singular vectors as an orthonormal basis, project the correction vector onto this subspace, and apply the projected vector uniformly to all 1,273 test questions. Results (Table~\ref{tab:subspace}) show that neither expanding rank nor changing the basis source closes the gap.

\section{Probe vs.\ Contrastive Vector Analysis}
\label{app:probe_analysis}

\begin{figure*}[t]
\centering
\includegraphics[width=\textwidth]{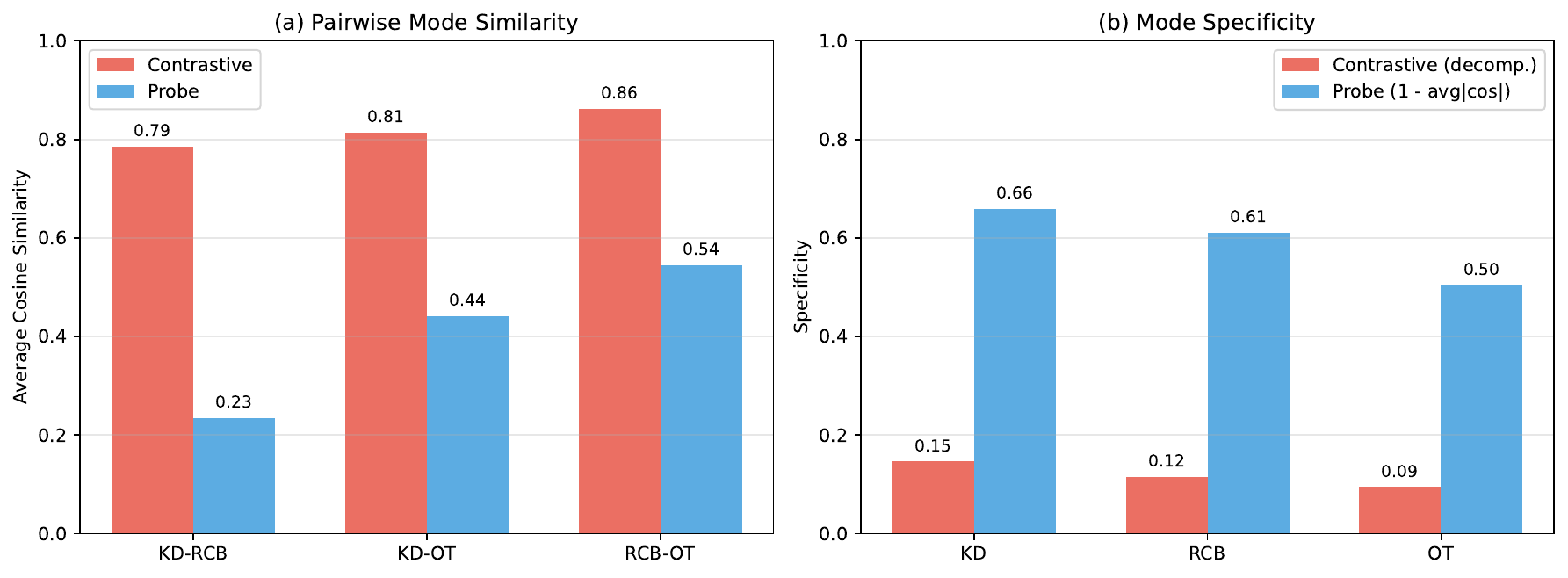}
\caption{Probe vs contrastive vector comparison. (a) Pairwise cosine similarity: probe vectors are much less correlated (0.23--0.54) than contrastive vectors (0.79--0.86), indicating the probe optimization finds more discriminative directions. (b) Mode specificity: probes achieve 5--6$\times$ higher specificity than contrastive decomposition.}
\label{fig:probe_vs_contrastive}
\end{figure*}

Per-layer logistic regression probes (PCA-50 features) reveal a strikingly different geometry from mean-difference contrastive vectors (Figure~\ref{fig:probe_vs_contrastive}). Probe weight vector pairs have average cosine 0.23--0.54, showing the probe finds more discriminative, mode-specific directions than the contrastive approach. Yet despite this geometric dissociation, both methods achieve comparable classification accuracy (probes: 51.2\% at layer 15; contrastive: 51.5\% at layer 16). This means multiple distinct directions in activation space support the same classification---further evidence that decodability does not identify a unique causal intervention target.

\section{Uncertainty Baseline Comparison}
\label{app:uncertainty}

\begin{table}[t]
\centering
\footnotesize
\setlength{\tabcolsep}{3pt}
\begin{tabular}{lcccc}
\toprule
Method & AUROC & 95\% CI & Acc@60\% & +pp \\
\midrule
Correctness probe (L21) & \textbf{0.716} & [0.699, 0.732] & 64.0 & +14.0 \\
\midrule
Hidden state norm (L21) & 0.530 & [0.511, 0.549] & 53.0 & +3.0 \\
Hidden state norm (L31) & 0.510 & [0.490, 0.528] & 52.1 & +2.1 \\
Logit margin (top1$-$top2) & 0.504 & [0.486, 0.523] & 51.8 & +1.8 \\
Max token probability & 0.505 & [0.486, 0.523] & 51.8 & +1.8 \\
Negative entropy & 0.505 & [0.487, 0.523] & 51.7 & +1.7 \\
\bottomrule
\end{tabular}
\caption{Uncertainty baseline comparison for selective abstention ($n = 3{,}600$ held-out traces, balanced 50\% correct / 50\% incorrect). All baselines are single-forward-pass metrics computed from the same hidden states. The correctness probe outperforms the best uncertainty baseline (hidden state norm) by $\Delta$AUROC = 0.186 ($p < 10^{-4}$, paired bootstrap, 5{,}000 resamples). AUROC is prevalence-independent; the ``+pp'' column shows gains relative to the 50\% balanced baseline at 60\% coverage.}
\label{tab:uncertainty}
\end{table}

\begin{table}[t]
\centering
\footnotesize
\setlength{\tabcolsep}{3pt}
\begin{tabular}{lcc}
\toprule
Method & AUROC & 95\% CI \\
\midrule
Correctness probe (L21) & \textbf{0.610} & [0.577, 0.642] \\
\midrule
Negative entropy & 0.569 & [0.534, 0.603] \\
Max token probability & 0.567 & [0.532, 0.601] \\
Logit margin (top1$-$top2) & 0.563 & [0.528, 0.596] \\
Hidden state norm (L31) & 0.506 & [0.472, 0.541] \\
Hidden state norm (L21) & 0.427 & [0.393, 0.462] \\
\bottomrule
\end{tabular}
\caption{Test-set uncertainty baselines ($n = 1{,}273$ test questions, temperature 0.1 generation). The correctness probe---trained on temperature 0.8 training traces---outperforms the best uncertainty baseline ($\Delta$AUROC = 0.041, $p = 0.009$, paired bootstrap). Baselines are computed directly on test-set logits and hidden states; the probe faces an additional generalization challenge (cross-temperature, cross-question transfer). CIs are 95\% bootstrap (5{,}000 resamples).}
\label{tab:uncertainty_test}
\end{table}

We focus on single-forward-pass baselines computed from the same hidden states for fair comparison; modern neural networks are known to be poorly calibrated \citep{guo2017calibration}, motivating hidden-state probes over raw confidence scores. Multi-sample methods such as semantic entropy \citep{kuhn2023semantic} require multiple generations, comparable in cost to self-consistency \citep{wang2023selfconsistency} (AUROC = 0.804), which we report as the multi-sample performance ceiling. Verbalized confidence \citep{xiong2024uncertainty} and $P(\text{True})$ probing \citep{kadavath2022language} remain untested single-pass alternatives.

\section{Concept Erasure Random Direction Control}
\label{app:random_control}

To test whether the mean-difference erasure damage (full test set: $\Delta = -3.6$pp, $p = 0.010$, $n = 1{,}273$) is specific to the failure-mode direction or a generic artifact of rank-1 projection, we compare the \OT{} direction against 10 random directions of equal rank on a 300-question subset under identical conditions. For each random direction $\mathbf{q}_{\text{rand}} \sim \mathcal{N}(0, \mathbf{I})$ (normalized), we construct $\mathbf{P}_{\text{rand}} = \mathbf{I} - \mathbf{q}_{\text{rand}}\mathbf{q}_{\text{rand}}^T$ and evaluate accuracy under the same generation protocol.

On $n = 300$ questions, the OT direction produces $\Delta_{\text{OT}} = -3.9$pp, while 10 random directions produce $\Delta_{\text{rand}} = +0.3 \pm 1.8$pp (mean $\pm$ SD; range $-2.0$ to $+3.3$pp). The OT-specific excess ($\Delta_{\text{OT}} - \bar{\Delta}_{\text{rand}} = -4.2$pp; OT falls below all 10 random directions) confirms that the erasure effect is direction-specific, not a generic consequence of removing any rank-1 subspace. Two interpretations are equally consistent: (1)~the failure-mode direction co-occurs with (shares subspace with) task-relevant computation at layer 16; or (2)~the OT direction is causally required for correct reasoning, and erasure removes information needed for both failure-mode encoding and successful computation. No random direction produces damage comparable to the \OT{} direction (worst random: $-2.0$pp vs.\ OT: $-3.9$pp).

\section{Probe-Gated Dynamic Steering}
\label{app:probe_gated}

We test whether dynamic, token-conditional steering can succeed where static interventions fail ($n = 300$). At each generation step, a correctness probe (trained on last-token hidden states at layer 16) evaluates $P(\text{correct})$ from the current token's hidden state. Steering is applied only when the probe signals low confidence.

\begin{table}[t]
\centering
\footnotesize
\setlength{\tabcolsep}{3pt}
\begin{tabular}{lcccc}
\toprule
Condition & Acc. & $\Delta$ & Corr. & Dmg. \\
\midrule
Baseline & 60.0 & --- & --- & --- \\
Static contrastive ($\alpha$=1.5) & 52.0 & $-$8.0 & 39 & 63 \\
\midrule
Probe-gated binary ($P < 0.5$) & 57.0 & $-$3.0 & 37 & 46 \\
Probe-gated scaled & 62.7 & +2.7 & 45 & 37 \\
Probe-gated threshold ($P < 0.3$) & 61.3 & +1.3 & 35 & 31 \\
Probe-gated probe-vec ($P < 0.5$) & 60.3 & +0.3 & 38 & 37 \\
\bottomrule
\end{tabular}
\caption{Probe-gated dynamic steering ($n = 300$). All values in \%; Corr.\ and Dmg.\ show counts (not \%). Static contrastive steering significantly damages accuracy ($p = 0.018$, McNemar). No dynamic gating condition produces reliable improvement (all McNemar $p > 0.3$). The probe reports $P(\text{correct}) < 0.5$ for $>$93\% of intermediate tokens, resulting in high steering rates; the strict threshold ($P < 0.3$) reduces this to 61\% but still yields $\Delta \approx 0$.}
\label{tab:probe_gated}
\end{table}

No dynamic condition reliably improves accuracy (Table~\ref{tab:probe_gated}; all McNemar $p > 0.3$). Static contrastive steering is significantly harmful ($\Delta = -8.0$pp, $p = 0.018$; 63 damages vs.\ 39 corrections). Dynamic gating reduces this damage---the best gating variants achieve $\Delta \approx 0$---but cannot push it into positive territory. A contributing factor is that the probe, trained on last-token hidden states, is poorly calibrated on intermediate generation tokens: it reports $P(\text{correct}) < 0.5$ for 93\% of tokens under binary gating. Even the conservative threshold ($P < 0.3$) steers 61\% of tokens. This highlights an additional challenge for dynamic steering: the probe's distributional assumptions break down during autoregressive generation, and the gating mechanism partially degenerates toward static application.

\section{Cross-Temperature Vector Comparison}
\label{app:cross_temp}

To directly test whether contrastive directions are preserved across the temperature gap (training at 0.8, evaluation at 0.1), we generate 5 traces at temperature 0.1 for 397 training questions (100 \OT{} + 100 \KD{} + 100 \RCB{} + 97 other), yielding 1,111 correct and 869 incorrect traces. We compute binary correct-vs-incorrect contrastive vectors at each layer and measure cosine similarity against the temperature 0.8 reference vectors.

At peak classification layers (14, 16, 17), the mean cosine is \textbf{0.413}; across mid-layers (10--24) the mean is 0.381; across all 32 layers it is 0.348. This indicates that contrastive directions are \emph{partially preserved} across temperatures---consistent with the probe's cross-temperature generalization (AUROC 0.610 on temperature 0.1 test data, Table~\ref{tab:uncertainty_test}) while explaining why absolute performance degrades relative to within-temperature evaluation (training AUROC 0.716).

We verify this is a genuine temperature effect, not a data artifact. First, the failure-mode composition at temperature 0.1 differs substantially (\KD{} contributes 68.8\% of incorrect traces vs.\ 33.3\% at temperature 0.8, as \OT{} questions become mostly correct under near-greedy decoding). However, reweighting the temperature 0.8 incorrect traces to match this composition yields cosine $>$0.98 with the original balanced vector, ruling out composition shift as an explanation. Second, subsampling the temperature 0.8 data to match the temperature 0.1 sample size (1,111 correct, 869 incorrect) yields cosine $>$0.99 with the full vector, ruling out estimation noise. Third, at temperature 0.1, 47.5\% of questions produce unanimous traces (all correct or all incorrect), raising the concern that the contrastive vector reflects question difficulty rather than reasoning quality. To test this, we decompose into a \emph{within-question} vector (from the 208 mixed-outcome questions only, where the same question produces both correct and incorrect traces) and a \emph{between-question} vector (contrasting all-correct vs.\ all-incorrect question means). Both show comparable alignment with the temperature 0.8 reference (within: 0.36--0.41; between: 0.36--0.37), indicating that the reduced cosine is not an artifact of between-question difficulty confounds but reflects a genuine geometric shift in how correctness is encoded at different temperatures.

\section{Steering Robustness Experiments}
\label{app:robustness}

We test supplementary robustness conditions. The main $n$=1,273 results (Section~\ref{sec:steering}) provide the primary evidence.

\paragraph{Same-temperature steering ($n = 300$).} To directly test whether the steering null results from train/eval temperature mismatch, we evaluate mode-specific contrastive steering ($\alpha = 1.5$) at the \emph{same} temperature used for annotation and vector construction (temperature 0.8) on a 300-question subset with 3 traces per condition (majority-vote accuracy). Steering at the matching temperature is \emph{more} harmful than at temperature 0.1: $\Delta = -6.7$pp (95\% CI $[-12.3, -1.0]$; McNemar $p = 0.031$; 29 corrections vs.\ 49 damages). Per-trace accuracy shows a similar pattern ($\Delta = -7.9$pp). This rules out temperature mismatch as an explanation for the steering null: the classification-correction gap persists---and widens---when the evaluation temperature matches the training distribution. The larger damage at temperature 0.8 likely reflects higher sampling variance amplifying the harmful perturbation.

\paragraph{Temperature sweep ($n = 100$).} We evaluate mode-specific contrastive steering at three additional temperature settings: greedy ($\Delta = +2.0$pp, baseline = 55.0\%), temperature 0.3 ($\Delta = -2.7$pp, baseline = 60.7\%), and temperature 0.7 ($\Delta = -1.3$pp, baseline = 56.7\%). All deltas fall within $[-2.7, +2.0]$pp, consistent with the $\Delta \approx 0$ pattern from the full-scale experiments at temperature 0.1.

\paragraph{Steering timing.} We compare full-sequence steering against early-only (first 50\% of tokens) and late-only (last 50\%) steering. Early-token steering is most harmful ($\Delta = -11.0$pp, $p = 0.022$), suggesting that perturbation during the critical initial reasoning phase actively disrupts generation. Full-sequence and late-only both produce $\Delta = -7.0$pp. These results are directionally consistent with the temporal mismatch analysis (Limitations): the failure-occurrence signal accumulates during generation, and static perturbation applied early---before the model commits to a reasoning path---causes the most damage.

\section{Complete Steering Configuration Sweep}
\label{app:alpha_sweep}

Table~\ref{tab:steering} and Table~\ref{tab:subspace} present representative configurations. For completeness, Table~\ref{tab:full_sweep} reports all tested configurations including omitted $\alpha$ values. All probe-guided experiments use $n = 1{,}175$ (questions with valid probe outputs); all others use $n = 1{,}273$.

\begin{table}[t]
\centering
\footnotesize
\setlength{\tabcolsep}{2.5pt}
\begin{tabular}{llcccc}
\toprule
Family & Config & $\alpha$ & Acc. & $\Delta$ & $p$ \\
\midrule
\multicolumn{6}{l}{\emph{Contrastive (n=1,273)}} \\
\; Uniform (shared) & L16 & 1.5 & 53.0 & $-$12.1 & $<$.001 \\
\; Mode-specific & L16 & 1.5 & 65.0 & $-$0.2 & .953 \\
\; Mode-specific & L16 & 3.0 & 65.3 & +0.2 & .952 \\
\midrule
\multicolumn{6}{l}{\emph{Probe-Guided (n=1,175)}} \\
\; Probe-uniform & L16 & 0.5 & 66.0 & $-$0.4 & .806 \\
\; Probe-uniform & L16 & 1.0 & 65.4 & $-$1.0 & .503 \\
\; Probe-uniform & L16 & 1.5 & 63.5 & $-$3.0 & .041 \\
\; Probe-mode & L16 & 0.5 & 64.8 & $-$1.7 & .233 \\
\; Probe-mode & L16 & 1.0 & 65.1 & $-$1.4 & .361 \\
\; Probe-mode & L16 & 1.5 & 65.3 & $-$1.2 & .418 \\
\midrule
\multicolumn{6}{l}{\emph{Multi-layer (n=1,273)}} \\
\; 3 layers & L13--17 & 1.5 & 64.6 & +0.2 & .956 \\
\; 5 layers & L11--19 & 1.5 & 62.8 & $-$1.6 & .280 \\
\midrule
\multicolumn{6}{l}{\emph{Prompt-End (n=1,273)}} \\
\; PE Uniform & L16 & 1.5 & 65.3 & +0.5 & .735 \\
\; PE Mode-specific & L16 & 1.5 & 66.4 & +1.6 & .241 \\
\; PE Mode-specific & L16 & 3.0 & 65.0 & +0.2 & .912 \\
\; PE Full composite & L16 & 1.5 & 65.7 & +0.9 & .542 \\
\midrule
\multicolumn{6}{l}{\emph{Strong Probe (n=1,273)}} \\
\; OT probe & L17 & 1.5 & 66.7 & +1.5 & .296 \\
\; OT probe & L17 & 3.0 & 61.4 & $-$3.8 & .010 \\
\; OT probe & L14 & 1.5 & 65.4 & +0.2 & .953 \\
\midrule
\multicolumn{6}{l}{\emph{Cross-architecture: Qwen (n=1,273)}} \\
\; Mode-specific & L18 & 0.5 & 61.8 & +0.2 & .827 \\
\; Mode-specific & L18 & 1.5 & 61.3 & $-$0.1 & 1.00 \\
\; Mode-specific & L18 & 2.0 & 61.9 & +0.3 & .779 \\
\; Mode-specific & L18 & 3.0 & 61.0 & $-$0.4 & .778 \\
\; Uniform & L18 & 1.5 & 61.9 & +0.5 & .677 \\
\; Uniform & L18 & 3.0 & 62.4 & +0.8 & .466 \\
\; Mode-specific & L5 & 1.5 & 61.7 & +0.1 & .942 \\
\; 3 layers & L16--18 & 1.5 & 61.3 & $-$0.3 & .770 \\
\; 5 layers & L14--18 & 1.5 & 60.6 & $-$0.9 & .403 \\
\bottomrule
\end{tabular}
\caption{Complete additive steering sweep on Llama-3.1-8B and Qwen2.5-7B (all $\alpha$ values tested). All values in \%. $\Delta$ in pp vs.\ run-matched baseline. McNemar $p$-values are two-sided. No configuration achieves significant improvement; probe-uniform $\alpha = 1.5$ is the only individually significant result and is harmful. The strong-probe L17/$\alpha$=3.0 is significantly \emph{harmful}. See Table~\ref{tab:subspace} for rank-$k$ subspace configurations (11 additional).}
\label{tab:full_sweep}
\end{table}

\section{Linear Accessibility Profile (LAP) Analysis}
\label{app:lap}

To independently diagnose why steering fails, we adapt the LAP framework of \citet{billa2026predicting}, which distinguishes three regimes: (1)~concept is output-aligned (logit-lens detects it, steering works); (2)~concept is nonlinearly encoded (probe detects, logit-lens fails); (3)~concept is not cleanly extractable (nothing works). We compute two metrics per layer:

\paragraph{A\textsubscript{lin} (logit-lens accuracy).} We project each hidden state onto the normalized \OT{} direction $\hat{\mathbf{d}}_{\text{OT}}^{(l)}$ and classify via optimal threshold, measuring how well the direction separates \OT{} vs.\ non-\OT{} in the output-aligned sense.

\paragraph{A\textsubscript{mlp} (probe accuracy).} The trained linear probe accuracy from Table~\ref{fig:layer_classification}b, serving as the upper bound on linearly decodable information.

\paragraph{Results.} Peak A\textsubscript{lin} = 58.6\% (layer 14), far below peak A\textsubscript{mlp} = 81.5\% (layer 17). The gap $\Delta = A_{\text{mlp}} - A_{\text{lin}} = 0.23$ across layers 5--25 indicates that \OT{} information is present in hidden states but \emph{not output-aligned}---the direction does not project to coherent vocabulary-space tokens. Examining the top-20 tokens from $\hat{\mathbf{d}}_{\text{OT}}^{(l)} \cdot \mathbf{W}_U^T$ at all layers reveals semantically incoherent token lists (e.g., layer 16: ``Knight,'' ``hra,'' ``iagnostics''; layer 24: ``eh,'' ``Engel,'' ``ser''), with 0/10 tokens semantically related to overthinking, correctness, or medical reasoning at the peak layer. This contrasts with steerable concepts (e.g., refusal), where logit-lens projections typically surface semantically coherent tokens (``sorry,'' ``cannot,'' ``harmful'').

\paragraph{Regime diagnosis.} The combination of high A\textsubscript{mlp} ($>$0.8) with low A\textsubscript{lin} ($\approx$ 0.59) places \OT{} in \textbf{Regime 2/3}: the concept is decodable by a trained probe but not accessible through the model's output pathway, consistent with the broader steering failure pattern. This provides an independent geometric explanation complementary to the specificity ratio analysis: the \OT{} direction does not align with the model's vocabulary projection, so additive perturbation along this direction produces incoherent logit shifts rather than targeted answer correction.

\end{document}